\documentclass[10pt,journal,compsoc]{IEEEtran}
\ifCLASSOPTIONcompsoc
  \usepackage[nocompress]{cite}
\else
  \usepackage{cite}
\fi
\ifCLASSINFOpdf
  \usepackage[pdftex]{graphicx}
  \graphicspath{{./}}
  \DeclareGraphicsExtensions{.pdf,.jpg,.jpeg,.png,}
\else
\fi
\usepackage{graphicx}
\usepackage{amsmath,amssymb} 
\usepackage{color}
\usepackage{multirow}
\usepackage{algorithm,algorithmic,bm,color}
\usepackage[colorlinks=true,allcolors=black]{hyperref}

\newcommand{\argmin}{\arg\!\min}
\newcommand{\st}{\mathrm{s.\,t.}}

\newcommand{\Cmat}{{\boldsymbol C}}
\newcommand{\Dmat}{{\boldsymbol D}}
\newcommand{\Emat}[0]{{{\boldsymbol E}}}

\newcommand{\Gmat}[0]{{{\boldsymbol G}}}

\newcommand{\Imat}{{\boldsymbol I}}

\newcommand{\Rmat}[0]{{{\boldsymbol R}}}

\newcommand{\Umat}{{{\boldsymbol U}}}
\newcommand{\Vmat}[0]{{{\boldsymbol V}}}

\newcommand{\Xmat}{{\boldsymbol X}}
\newcommand{\Ymat}[0]{{{\boldsymbol Y}}}
\newcommand{\Zmat}{{\boldsymbol Z}}

\newcommand{\bv}{\boldsymbol{b}}
\newcommand{\cv}{{\boldsymbol{c}}}

\newcommand{\ev}[0]{{\boldsymbol{e}}}

\newcommand{\gv}[0]{{\boldsymbol{g}}}

\newcommand{\qv}[0]{{\boldsymbol{q}}}
\newcommand{\rv}[0]{{\boldsymbol{r}}}

\newcommand{\wv}{\boldsymbol{w}}

\newcommand{\xv}{\boldsymbol{x}}
\newcommand{\yv}{\boldsymbol{y}}

\newcommand{\Sigmamat}{\boldsymbol{\Sigma}}

\newcommand{\Phimat}{\boldsymbol{\Phi}}

\newcommand{\thetav}{\boldsymbol{\theta}}

\newcommand{\tsp}{^{\mathsf{T}}}
\newcommand{\inv}{^{-1}}
\newcommand{\ie}{{\em i.e.}}
\newcommand{\eg}{{\em e.g.}}
\newcommand{\wrt}{{\em w.r.t.\,}}

\newtheorem{theorem}{Theorem}

\begin{document}
%
\title{Rank Minimization for\\ Snapshot Compressive Imaging}
%
%
%
%

\author{
        Yang~Liu,
        Xin~Yuan,~\IEEEmembership{Senior~Member,~IEEE,}
        Jinli~Suo,
        David~J.~Brady,~\IEEEmembership{Fellow,~IEEE,}
        and~Qionghai~Dai,~\IEEEmembership{Senior~Member,~IEEE}
\IEEEcompsocitemizethanks{\IEEEcompsocthanksitem Y.~Liu, J.~Suo and Q.~Dai are with the Department of Automation, Tsinghua University, Beijing 100084, China.\protect\\
E-mails: y-liu16@mails.tsinghua.edu.cn, jlsuo@tsinghua.edu.cn, qhdai@tsinghua.edu.cn.
\IEEEcompsocthanksitem X.~Yuan is with Nokia Bell Labs, Murray Hill, New Jersey, 07974, USA.\protect\\
E-mail: xyuan@bell-labs.com.
\IEEEcompsocthanksitem D.~J.~Brady is with the Department of Electrical and Computer Engineering,
Duke University, Durham, NC, 27708, USA. \protect\\ E-mail:
david.brady@duke.edu.
\IEEEcompsocthanksitem Corresponding author: Jinli~Suo.
\IEEEcompsocthanksitem Y. Liu and X. Yuan contribute equally to this paper.
}
\thanks{Manuscript updated \today.}
}

\IEEEtitleabstractindextext{%
\begin{abstract}
Snapshot compressive imaging (SCI) refers to compressive imaging systems where multiple frames are mapped into a single measurement,
with video compressive imaging and hyperspectral compressive imaging as two representative applications. 
Though exciting results of high-speed videos and hyperspectral images have been demonstrated, the poor reconstruction quality precludes SCI from wide applications. 
This paper aims to boost the reconstruction quality of SCI via exploiting the high-dimensional structure in the desired signal.  
We build a joint model to integrate the nonlocal self-similarity of video/hyperspectral frames and the rank minimization approach with the SCI sensing process.
Following this, an alternating minimization algorithm is developed to solve this non-convex problem.
We further investigate the special structure of the sampling process in SCI to tackle the computational workload and memory issues in SCI reconstruction.
Both simulation and real data (captured by four different SCI cameras) results demonstrate that our proposed algorithm leads to significant improvements compared with current state-of-the-art algorithms.  We hope our results  will encourage the researchers and engineers
to pursue further in compressive imaging for real applications. 
\end{abstract}

\begin{IEEEkeywords}
Compressive sensing, computational imaging, coded aperture, image processing, video processing, nuclear norm, rank minimization, low rank, hyperspectral images, coded aperture snapshot spectral imaging (CASSI), coded aperture compressive temporal imaging (CACTI).
\end{IEEEkeywords}}

\maketitle

\IEEEdisplaynontitleabstractindextext

%
\IEEEpeerreviewmaketitle

\IEEEraisesectionheading{\section{Introduction}\label{sec:introduction}}

%
%
%
%
\IEEEPARstart{C}{ompressive} sensing (CS) \cite{Donoho06ITT,Candes06ITT} has inspired practical {compressive imaging} systems to capture high-dimensional data such as videos~\cite{Hitomi11ICCV,Reddy11CVPR,Patrick13OE,Yuan14CVPR,Sun16OE,Sun17OE} and hyper-spectral images~\cite{Gehm07,Wagadarikar08CASSI,Wagadarikar09CASSI,Yuan15JSTSP,Cao16SPM}.
In video CS, high-speed frames are modulated at a higher frequency than the capture rate of the camera, which is working at a low frame rate.
In order to achieve ultra-high frame rate~\cite{Gao14_Nature}, for multiple frames, there is  only a single measurement  available per pixel in these high-dimensional compressive imaging systems. 
In this manner, each captured measurement frame can recover a number of high-speed frames, which is dependent on the coding strategy, \eg, 148 frames reconstructed from a snapshot in~\cite{Patrick13OE};  
if the CS imaging system is working (sampling measurements) at 30 frames per second (fps), we can achieve a video frame rate of higher than 4,000 fps.
In hyper-spectral image CS, the wavelength dependent coding is implemented by a coded aperture (physical mask) and a disperser~\cite{Wagadarikar08CASSI,Wagadarikar09CASSI}. More than 30 hyperspectral images have been reconstructed from a snapshot measurement.  
These systems can be recognized as {\em snapshot compressive imaging} (SCI) systems.

Though these SCI systems have led to exciting results, the poor quality of reconstructed images precludes them from wide applications. Therefore, algorithms with high quality reconstruction are desired.
This gap will be filled by our paper via exploiting the performance of optimization based reconstruction algorithms.    
It is worth noting that different from traditional CS~\cite{Duarte08SPM} using random sensing matrices, 
in SCI, the sensing matrix has special structures and it is not random.
Though this has raised the challenge in theoretical analysis~\cite{Jalali18ISIT}, we can gain the speed in the algorithmic development (details in Sec.~\ref{Sec:ADMM}) by using this structure. 
Most importantly, this structure is inherent in the hardware development of SCI, such as the video CS~\cite{Patrick13OE} and spectral CS~\cite{Wagadarikar09CASSI}. 


\subsection{Motivation}
%
%
%
\begin{figure*}[htbp!]
	\centering
	{\includegraphics[width=2\columnwidth]{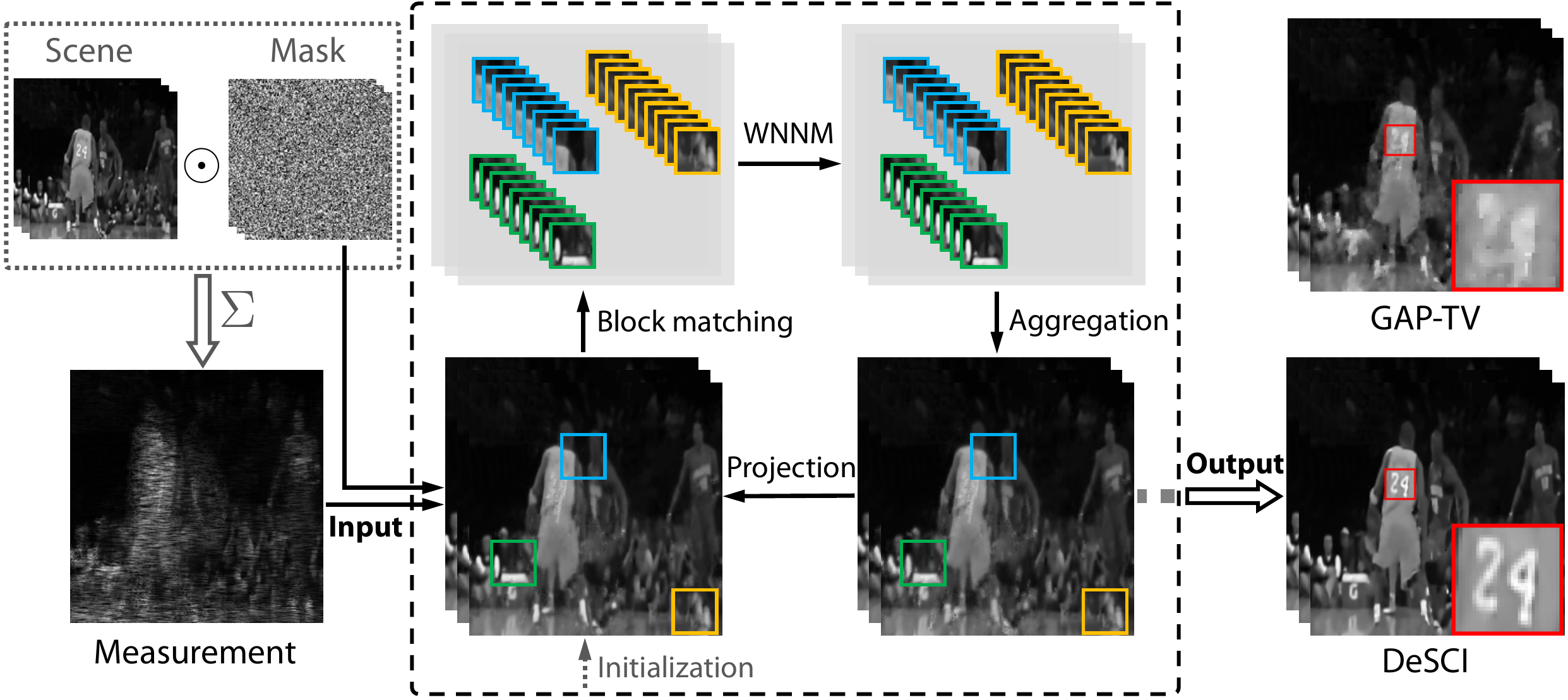}}
	\caption{ Flowchart of our proposed algorithm (DeSCI) for SCI reconstruction. Left: the sensing (compressively sampling) process of video SCI~\cite{Patrick13OE}. Middle: The proposed rank minimization based reconstruction algorithm, where the projection and WNNM for patch groups are iteratively performed.
		Right:  Our reconstruction result and that of the GAP-TV method~\cite{Yuan16ICIP_GAP} are shown for comparison in the upper part.}
	\label{fig:flowchart}
\end{figure*}
Without loss of generality, we use video CS as an example below to describe the motivation and contribution of this work. Hardware review and results of hyperspectral images are presented along with videos in corresponding sections. 
Different from traditional CS~\cite{Duarte08SPM,Yuan16SJ}, in SCI, the desired signal usually lies in high dimensions; for instance, a 148-frame video clip with spatial resolution $256\times256$ (pixels) is recovered from a single $256\times256$ (pixels) measurement frame.
This paper aims to address the following key challenges in SCI reconstruction.
\begin{itemize}
	\item [1)] How to exploit the structure information in high-dimensional videos and hyperspectral images?
	\item [2)] Are there some approaches for other (image/video processing) tasks that can be used in SCI to improve the  reconstruction quality? 
\end{itemize}

One of the state-of-the-art algorithms for SCI, the Gaussian mixture model (GMM) based algorithms~\cite{Yang14GMM,Yang14GMMonline} only exploited the sparsity of the video patches. 
In widely used videos (\eg, the \texttt{Kobe} dataset in Fig.~\ref{fig:gmm_gaptv_desci_frames}), these GMM based algorithms cannot provide reconstructed video frames with a PSNR (peak-signal-to-noise-ratio) more than 30dB. Similar to simulation, for real data captured by SCI cameras, the results of GMM suffer from blurry and other unpleasant artifacts (Fig.~\ref{sfig:real_chopperwheel}).
While the blurry is mainly due to the limitation of sparse priors used in GMM, the unpleasant artifacts might be due to the system noise.  
Motivated by these limitations of GMM, to develop a better reconstruction algorithm, high-dimensional structure information is necessary to be investigated, for example, the nonlocal similarity across temporal and spectral domain in addition to the spatial domain. 
Moreover, an algorithm which is robust to noise is highly demanded as the system noise is unavoidable in real SCI systems.

A reconstruction framework that takes advantage of more comprehensive structural information can potentially outperform existing algorithms. One recent proposal in CS is to use algorithms that are designed for other data
processing tasks such as denoising~\cite{Mertzler14Denoising}, which employed advanced denoising algorithms in the approximate message passing framework to achieve state-of-the-art results in image CS. 
This motivates us to develop advanced reconstruction algorithms for SCI by leveraging the advanced techniques in video denoising~\cite{Ji10CVPR}. 
Meanwhile, recent researches on rank minimization have led to significant advances in other image and video processing tasks~\cite{Ji10CVPR,Cai10SIAM,Gu14CVPR,Gu17IJCV,Liu10ICML}.
These advances have great potentials to boost the performance of reconstruction algorithms for SCI and this will be investigated in our paper.

\subsection{Contributions}
While the rank minimization approaches have been investigated extensively for image processing, extending them to the video and hyperspectral image cases, especially the SCI is nontrivial. 
Particularly, to achieve high-speed videos, in SCI, the measurement is not video frames, but the linear combination of a number of frames. In this manner, the rank minimization methods cannot be imposed directly as used in image processing.
To conquer this challenge, a new reconstruction framework is proposed, which incorporates the rank minimization as an intermediate step during reconstruction.
{Specifically, by integrating the compressive sampling model in SCI with the weighted nuclear norm minimization (WNNM)~\cite{Gu14CVPR} for video patch groups (see details in Sec.~\ref{Sec:Problem}), a joint model for SCI reconstruction is formulated.} 
To solve this problem, the alternating direction method of multipliers (ADMM)~\cite{Boyd11ADMM} method is employed to develop an iterative optimization algorithm for SCI reconstruction.

Fig.~\ref{fig:flowchart} depicts the flowchart of our proposed algorithm for SCI reconstruction. {After the measurement is captured by the SCI systems (left part in Fig.~\ref{fig:flowchart}), our proposed algorithm, dubbed DeSCI (decompress SCI, middle part in Fig.~\ref{fig:flowchart}), performs the projection (which projects the measurement/residual to the signal space to fit the sampling process in the SCI system) and WNNM denoising (imposing signal structural priors) for video patch groups (with details in Sec.~\ref{Sec:Problem}) iteratively}.
In the right part, we show the reconstruction results of our proposed DeSCI algorithm on the \texttt{Kobe} data used in~\cite{Yang14GMM} and the results  of the GAP-TV (generalized alternating projection based total variation) method~\cite{Yuan16ICIP_GAP} are shown for comparison in the upper part. It can be seen clearly that our DeSCI algorithm provides better image/video quality than GAP-TV.

Moreover, our proposed DeSCI algorithm has boosted the reconstruction quality of real data captured by four different SCI cameras, \ie, CACTI (coded aperture compressive temporal imaging)~\cite{Patrick13OE}, color-CACTI~\cite{Yuan14CVPR}, CASSI (coded aperture snapshot spectral imaging)~\cite{Wagadarikar09CASSI} and the high-speed stereo camera~\cite{Sun17OE}; please refer to Fig~\ref{sfig:real_chopperwheel} to Fig.~\ref{sfig:real_hsi_object_all} for visual comparisons. These real data results clearly demonstrate the advantages of our proposed algorithms, \eg, robust to noise. 
We thus strongly believe that
our findings have significant practical values for both research and applications.
We hope our encouraging results will inspire researchers and engineers
to pursue further in compressive imaging.

\subsection{Related work and organization of this paper}
As SCI reconstruction is an ill-posed problem, to solve it, different priors have been employed, which can be categorized into total variation (TV)~\cite{Yuan16ICIP_GAP}, sparsity in transform domains~\cite{Reddy11CVPR,Patrick13OE,Yuan14CVPR}, sparsity in over-complete dictionaries~\cite{Hitomi11ICCV,Yuan15JSTSP}, and the GMM based algorithms~\cite{Yang14GMM,Yang14GMMonline}. 
The group sparsity based algorithms~\cite{Maggioni2012VideoDD} and the nonlocal self-similarity~\cite{Dong14TIP} model, which has led to significant advances in image processing, have not been used in these SCI reconstruction algorithms. This is investigated in our paper. 
Most recently, the deep learning techniques have been utilized for video CS~\cite{Iliadis18DSPvideoCS,2016arXivVideoCS}.
We are not aiming to compete with these algorithms as they are usually complicated and require data to train the neural networks. 
Furthermore, some of these algorithms require the sensing matrix being spatially repetitive~\cite{Iliadis18DSPvideoCS}, which is very challenging (or unrealistic) in real applications. In addition, we have noticed that limited improvement (around 2dB) has been obtained using these deep learning techniques in~\cite{Iliadis18DSPvideoCS} compared with GMM. By contrast, our proposed algorithm has improved the results significantly ($>4$dB) over GMM.
Specifically, we employ rank minimization approach to the nonlocal similar patches in videos and hyperspectral images. In this manner, the reconstruction results have been improved dramatically and this would pave the way of wide applications of SCI systems, such as high-speed videos~\cite{Yuan14CVPR}, hyperspectral images~\cite{Yuan15JSTSP} and three-dimensional high-accuracy high-speed in-door localizations~\cite{Sun17OE}. 

The rest of this paper is organized as follows.
Sec.~\ref{sec:review_of_real_snapshot_compressive_imagers} reviews the two standard SCI systems, namely, the video SCI and hyperspectral image SCI systems.
The mathematical model of SCI is introduced in Sec.~\ref{sec:math_model}.
Sec.~\ref{Sec:Algo} develops the rank minimization based algorithm for SCI reconstruction.
Extensive simulation results are presented in Sec.~\ref{Sec:Sim_Results} to demonstrate the efficacy of the proposed algorithm, and real data results are shown in Sec.~\ref{Sec:Real_Results}.
Sec.~\ref{Sec:Dis} concludes the paper and discusses future research directions.

\section{Review of snapshot compressive imaging systems} 
\label{sec:review_of_real_snapshot_compressive_imagers}
The last decade has seen a number of SCI systems~\cite{Gehm07,Wagadarikar08CASSI,Hitomi11ICCV,Reddy11CVPR,Patrick13OE,Kshitij13ToG,Gao14_Nature,Lin14ToG,Yuan14CVPR,Brady15AOP,Tsai15OL,Sun17OE,Brady18Optica} with the development of compressive sensing~\cite{Donoho06ITT,cs_Candes06stable,cs_Candes06}. The underlying principle is encoding the high-dimensional data on a 2D sensor with dispersion for spectral imaging~\cite{Gehm07,Wagadarikar08CASSI,Lin14ToG}, temporal-variant mask for high-speed imaging~\cite{Hitomi11ICCV,Reddy11CVPR,Patrick13OE,Gao14_Nature}, and angular variation for light-field imaging~\cite{Kshitij13ToG}. Recently, several variants explore more than three dimensions of the scene~\cite{Yuan14CVPR,Tsai15OL,Sun17OE,Wang17TPAMI,Wang18TPAMI}, which paves the way for plenoptic imaging~\cite{Adelson1991plenoptic}. 

We validate the proposed DeSCI methods on two typical snapshot compressive imaging systems, \ie, snapshot compressive imagers such as the CACTI system~\cite{Patrick13OE} and the CASSI system~\cite{Gehm07,Wagadarikar08CASSI}, as shown in Fig.~\ref{sfig:cacti} and Fig.~\ref{sfig:cassi}, respectively. {Similar approaches could be adapted for other compressive imaging systems with minor modifications, since we only need to change the sensing matrix for various coding strategies and the nonlocal self-similarity always holds for natural scenes.}

\subsection{Snapshot-video compressive imaging system} 
\label{sub:snapshot_video_compressive_imaging_system}
\begin{figure}
	\centering
	{\includegraphics[width=1\columnwidth]{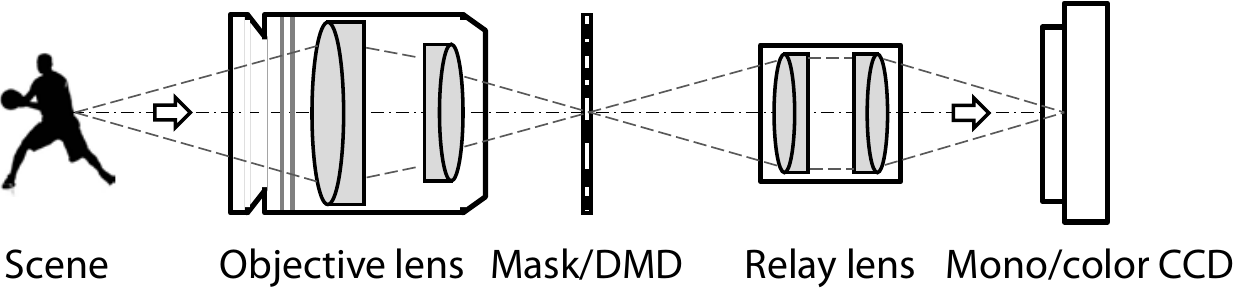}}
	\caption{Schematic of the coded aperture compressive temporal imaging (CACTI) system~\cite{Patrick13OE}. A snapshot on the CCD encodes tens of temporal frames of the scene coded by the spatial-variant mask, \eg, the shifting mask or different patterns on the digital micromirror device (DMD). The mask/DMD and the mono/color detector, \ie, CCD, are in the conjugate image plane of the scene.}
	\label{sfig:cacti}
\end{figure}

In video snapshot compressive imagers, \ie, the CACTI system~\cite{Patrick13OE}, the high-speed scene is collected by the objective lens and spatially coded by the temporal-variant mask, such as the shifting mask~\cite{Patrick13OE} or different patterns on the digital micromirror device (DMD) or the spatial light modulator (SLM)~\cite{Hitomi11ICCV,Reddy11CVPR,Sun16OE}, as shown in Fig.~\ref{sfig:cacti}. Then the coded scene is detected by the monochrome or color CCD (Charge-Coupled Device) for grayscale~\cite{Patrick13OE} and color~\cite{Yuan14CVPR} video capturing, respectively. A snapshot on the CCD encodes tens of temporal frames of the high-speed scene. The number of coded frames for a snapshot is determined by the number of variant codes (of the mask) within the integration time. 

\subsection{Snapshot-spectral compressive imaging system} 
\label{sub:snapshot_spectral_compressive_imaging_system}
\begin{figure}
	\centering
	{\includegraphics[width=1\columnwidth]{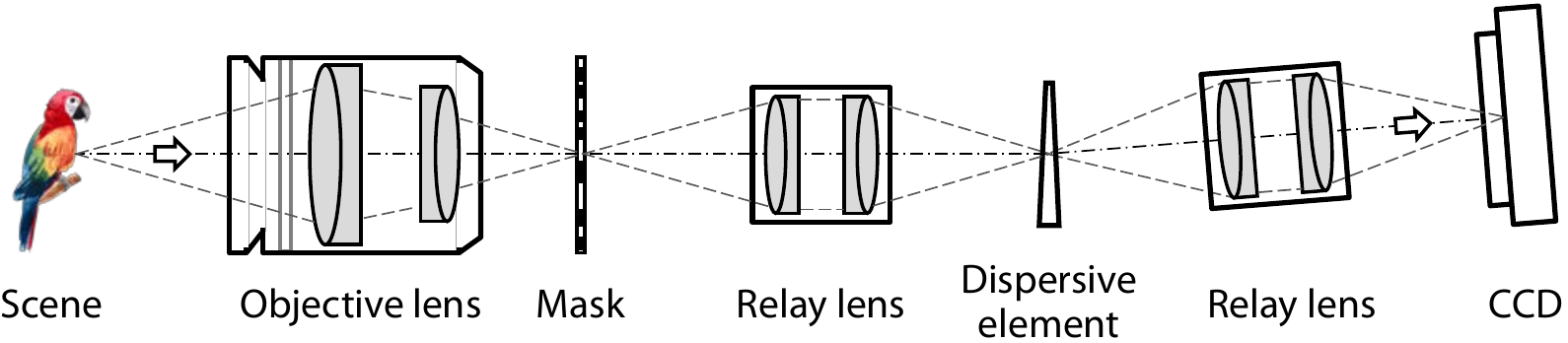}}
	\caption{Schematic of the coded aperture snapshot spectral imaging (CASSI) system~\cite{Wagadarikar08CASSI}. A snapshot on the CCD encodes tens of spectral bands of the scene spatially coded by the mask and spectrally coded by the dispersive element. The mask, dispersive element and the CCD are in the conjugate image plane of the scene.}
	\label{sfig:cassi}
\end{figure}

In spectral snapshot compressive imagers, \ie, the CASSI system~\cite{Gehm07,Wagadarikar08CASSI}, the spectral scene is collected by the objective lens and spatially coded by a fixed mask, as shown in Fig.~\ref{sfig:cassi}. Then the coded scene is spectrally dispersed by the dispersive element, such as the prism or the grating. The spatial-spectral coded scene is detected by the CCD. A snapshot on the CCD encodes tens of spectral bands of the scene. The number of coded frames for a snapshot is determined by the dispersion property of the dispersive element and the pixel size of the mask and the CCD~\cite{Wagadarikar08CASSI}.

\section{Mathematical model of snapshot compressive imaging}
\label{sec:math_model}
Mathematically, the measurement in the SCI systems can be  modeled by~\cite{Patrick13OE}
\begin{equation} \label{Eq:yPhix}
  {\yv = \Phimat \xv + \gv}\,,
\end{equation}
where $\Phimat \in {\mathbb R}^{n\times nB}$ is the sensing matrix, $\xv\in {\mathbb R}^{nB}$ is the desired signal, and $\gv\in {\mathbb R}^{n}$ denotes the noise.
Unlike traditional CS, the sensing matrix considered here is not a dense matrix. 
In SCI, {\eg, video CS as in CACTI~\cite{Patrick13OE,Yuan14CVPR} and spectral CS as in CASSI~\cite{Wagadarikar09CASSI},} the matrix $\Phimat$ has a very specific structure and can be written as
\begin{equation} \label{Eq:Hmat_strucutre}
  {\Phimat = \left[\Dmat_1,\dots, \Dmat_B\right]}\,,
\end{equation}
where $\{\Dmat_k\}_{k=1}^B$ are diagonal matrices. 

Taking the SCI in CACTI~\cite{Patrick13OE} as an example, consider that $B$ high-speed frames $\{\Xmat_k\}_{k=1}^B \in {\mathbb R}^{n_x \times n_y}$ are modulated by the masks $\{\Cmat_k\}_{k=1}^B \in {\mathbb R}^{n_x \times n_y}$, correspondingly (Fig.~\ref{sfig:cacti}). The measurement $\Ymat \in {\mathbb R}^{n_x \times n_y}$ is given by
\begin{equation} \label{Eq:YXC}
  { \Ymat = \sum_{k=1}^B \Xmat_k \odot  \Cmat_k + \Gmat}\,, 
\end{equation}
where $\odot$ denotes the Hadamard (element-wise) product.
For all $B$ pixels (in the $B$ frames) at the position  $(i,j)$, $i = 1,\dots, n_x$; $j = 1,\dots, n_y$, they are collapsed to form one pixel in the measurement (in one shot) as
\begin{equation}
  { y_{i,j} = \sum_{k=1}^B c_{i,j,k} x_{i,j,k} + g_{i,j}}\,.
\end{equation}
By defining 
\begin{equation} \label{Eq:xv1toB}
  {\xv = \left[
	\xv_1\tsp,\dots,\xv_B\tsp
	\right]\tsp}\,,
\end{equation}
where $\xv_k = {\rm vec}(\Xmat_k)$,  and $\Dmat_k = {\rm diag}({\rm vec}(\Cmat_k))$, for $ k=1,\dots, B$, we have the vector formulation of Eq.~\eqref{Eq:yPhix}, where $n = n_xn_y$.
Therefore, $\xv \in {\mathbb R}^{n_xn_yB}$, $\Phimat\in {\mathbb R}^{n_xn_y \times (n_xn_yB)}$, and the {\em compressive sampling rate in  SCI is equal to  $1/B$}.
\footnote{Multiple measurements have also been investigated in~\cite{Kittle10AO,Arguello13TIP} for hyperspectral image CS, and our algorithm can be used in that system with minor modifications. The code design has also been investigated in~\cite{Arguello13TIP} for multiple shot  hyperspectral image CS. However, this is out of the scope of this paper.}
It is worth noting that due to the special structure of $\Phimat$ in \eqref{Eq:Hmat_strucutre}, we have $\Phimat\Phimat\tsp$ being a diagonal matrix. This fact will be  useful to derive the efficient algorithm in Sec.~\ref{Sec:ADMM} for handling the massive data in SCI.

One natural question is that is it {\em theoretically} possible to recover $\xv$ from the measurement $\yv$ defined in Eq.~\eqref{Eq:yPhix}, for $B>1$? Most recently, this has been addressed in~\cite{Jalali18ISIT} by using the compression-based compressive sensing regime~\cite{jalali2016compression} via the following theorem, where $\{f,g\}$ denotes the encoder/decoder, respectively.
\begin{theorem}\label{thm:main-csp}
	\cite{Jalali18ISIT} Assume that $\forall \xv\in{\cal Q}$, $\|\xv\|_{\infty}\leq {\frac{\rho}{2}}$. Further assume  the rate-$r$ code achieves distortion $\delta$ on $\cal{Q}$.  
	Moreover, for $i=1,\ldots,B$, $\Dmat_i={\rm diag}(D_{i1},\ldots,D_{in})$, and $\{D_{ij}\}_{j=1}^{n}\stackrel{\rm i.i.d.}{\sim} {\cal N}(0,1)$. For $\xv\in{\cal Q}$ and $\yv=\sum_{i=1}^B\Dmat_i\xv_i,$ let $\hat{\xv}$ denote the solution of compressible signal pursuit optimization.  Assume that $\epsilon>0$ is a free parameter, such that  $\epsilon \leq {\frac{16}{3}}$. Then,
	\begin{align}
	{ {\frac{1}{nB}}\|\xv-\hat{\xv}\|_2^2\leq \delta +\rho^2\epsilon},
	\end{align}
	with a probability larger than $	 { 1- 2^{nBr+1} {\rm e}^{- n(\frac{3\epsilon}{32})^2}}$.
	%
\end{theorem}

Details of the compressible signal pursuit optimization and the proof can be found in~\cite{Jalali18ISIT}. Most importantly, Theorem~\ref{thm:main-csp} characterizes the performance of SCI recovery by connecting the parameters of
the (compression/decompression) code, its rate $r$ and its distortion $\delta$, to the number of frames
$B$ and the reconstruction quality.
This theoretical finding strongly encourages our algorithmic design for SCI systems.

\section{Rank minimization for signal reconstruction in snapshot compressive imaging \label{Sec:Algo}}
In this section, we first briefly review the rank minimization algorithms and the joint model is developed in Sec.~\ref{Sec:Problem}. The proposed algorithm is derived in Sec.~\ref{Sec:ADMM}.
\subsection{Rank minimization\label{Sec:RankM}}
In image/video processing, since the matrix formed by nonlocal similar patches in a natural image is of low rank, a series of low-rank matrix approximation methods have been proposed for various tasks~\cite{Gu14CVPR,Peng12PAMI,CandesRPCA,Candes09LR}. 
The main goal of low-rank matrix approximation is to recover the underlying low-rank structure of a matrix from its degraded/corrupted observation. 
Within these frameworks, nuclear norm minimization (NNM) \cite{Cai10SIAM} is the most representative one and this will be used in our work.
Specifically, given a data matrix $\textbf{\emph{X}}\in\mathbb{R}^{b\times m}$, the goal of NNM is to find a low-rank matrix $\textbf{\emph{Z}}\in\mathbb{R}^{b\times m}$ of rank $r \leq \min\{{b, m}\}$, which satisfies the following objective function,
\begin{equation}
  {\hat{\textbf{\emph{Z}}}=\argmin_{\textbf{\emph{Z}}}\,\frac{1}{2}\|\textbf{\emph{X}}-\textbf{\emph{Z}}\|_\mathrm{F}^2 +\lambda\|\textbf{\emph{Z}}\|_\ast\,,}
\label{eq:1}
\end{equation} 
where $\|\cdot\|_\ast$ is the nuclear norm, \ie, the sum of the singular values of a given matrix, $\|\cdot\|_\mathrm{F}$ denotes the Frobenius norm and $\lambda>0$ is the regularization parameter. 
Despite the theoretical guarantee of the singular value thresholding model \cite{Cai10SIAM}, it has been observed that the recovery performance of such a convex relaxation will degrade in the presence of noise, and the solution can seriously deviate from the original solution of the rank minimization problem \cite{Xie16TIP_spnorm}. To mitigate this issue, Gu $\emph{et al}.$ \cite{Gu14CVPR} proposed the weighted nuclear norm minimization (WNNM) model, which is essentially the reweighted $\ell_1$-norm \cite{Candes08weightedL1} of the singular values of the desired matrix, and WNNM has led to state-of-the-art image denoising results. 
Though generally the WNNM problem is nonconvex, for non-descendingly ordered weights as used in this paper, we can get the optimal solution in closed-form by the weighted soft-thresholding operator~\cite{Gu17IJCV}. 
In the following, we will investigate how to use WNNM in the SCI systems.


\subsection{Integrating WNNM to SCI \label{Sec:Problem}}
To be concrete, all video frames $\{\Xmat_k\}_{k=1}^B$ are divided into $N$ overlapping patches of size $\sqrt{d}\times \sqrt{d}$, and each patch is denoted by a vector ${\textbf{\emph{z}}}_i\in\mathbb{R}^d, i=1,2,...,N$. For the $i^{th}$ patch $\textbf{\emph{z}}_i$,  its $M$ similar patches are selected from a surrounding (searching) window with $L \times L \times H$ pixels to form a set ${\cal S}_i$, where $L\times L$ denotes the window size in space and $H$ signifies the window size in time (across frames). 
After this, these patches in ${\cal S}_i$ are stacked into a matrix $\Zmat_i\in\mathbb{R}^{d\times M}$, \ie,
\begin{equation}
  {\Zmat_i=[{\textbf{\emph{z}}}_{i,1}, {\textbf{\emph{z}}}_{i,2},...,{\textbf{\emph{z}}}_{i,M}]\,.}
\label{eq:4}
\end{equation}
This matrix $\Zmat_i$ consisting of patches with similar structures is thus called a group, where $\{{\textbf{\emph{z}}}_{i,j}\}_{j=1}^M$ denotes the $j^{th}$ patch in the $i^{th}$ group. 
Since all patches in each data matrix have similar structures, the constructed data matrix ${\textbf{\emph{Z}}}_i$ is of {\em low rank}. 

By using this rank minimization as a constraint, the SCI problem in \eqref{Eq:yPhix}
can be formulated as
\begin{eqnarray} \label{eq:VCS_NNM}
  {{\hat \xv} = \argmin_{\xv} \frac{1}{2}\|\yv -\Phimat\xv\|_2^2 + \lambda \sum_i \|\Zmat_i\|_\ast}\,, 
\end{eqnarray}
where $\lambda$ is a parameter to balance these two terms, and recall that $\{\Zmat_i\}_{i=1}^N$ is constructed from $\xv$.

Given the fact that WNNM~\cite{Gu14CVPR} provides better results than NNM, we modify the model in~\eqref{eq:VCS_NNM} to
\begin{eqnarray} \label{eq:VCS_WNNM}
  {{\hat \xv} = \argmin_{\xv} \frac{1}{2}\|\yv -\Phimat\xv\|_2^2 + \lambda \sum_i \|\Zmat_i\|_{w,*}\,.}
\end{eqnarray}
Here \begin{equation}
  {\|\Zmat_i\|_{w,*} = \sum_{j=1}^{{\rm min}\{d,M\}} w_j \sigma_j}
\end{equation}
is the weighted nuclear norm of the matrix $\Zmat_i$ with weights defined in the vector $\wv = \{w_1, \dots, w_{{\rm min}(d,M)}\}$, where $w_j\ge0$ is the weight assigned to $\sigma_j$, the $j^{th}$ singular value of $\Zmat_i$.
The model in \eqref{eq:VCS_WNNM} is denoted as SCI-WNNM below.

\subsection{Solving the SCI-WNNM problem \label{Sec:ADMM}}
Under the ADMM~\cite{Boyd11ADMM} framework,  
we introduce an auxiliary variable $\thetav$ to the problem in \eqref{eq:VCS_WNNM}, 
\begin{eqnarray} \label{eq:Admm_WNNM_1}
  {{\hat \xv} = \argmin_{\xv} \frac{1}{2}\|\yv -\Phimat\thetav\|_2^2 + \lambda \sum_i \|\Zmat_i\|_{w,*}\,, ~\,\st~ \xv = \thetav}\,,
\end{eqnarray} 
where again $\{\Zmat_i\}_{i=1}^N$ is constructed from $\xv$.
Eq.~\eqref{eq:Admm_WNNM_1} can be translated into three sub-problems:
\begin{align}
\thetav^{(t+1)} &=  {\argmin_{\thetav}\frac{1}{2}\|\yv-\Phimat\thetav\|_2^2 + \frac{\gamma}{2} \|\thetav - \xv^{(t)} - \bv^{(t)}\|_2^2}\,, \label{Eq:theta_sub}\\
\xv^{(t+1)} &=   {\argmin_{\xv} \lambda \sum_i \|\Zmat_i\|_{w,*} + \frac{\gamma}{2} \|\thetav^{(t+1)} - \xv - \bv^{(t)}\|_2^2}\,, \label{Eq:x_sub}\\
\bv^{(t+1)} &= \bv^{(t)} - (\thetav^{(t+1)} - \xv^{(t+1)})\,. \label{Eq:ADMM_b1}
\end{align}
We derive the solutions to these sub-problems below, and without confusion, we discard the iteration index $t$.


\textbf{Solve $\thetav$:}
Given $\{\xv, \bv,\Phimat, \yv\}$, Eq~\eqref{Eq:theta_sub} is a quadratic form and has a closed-form solution
\begin{eqnarray}\label{Eq:ADMM_theta1}
  {\thetav = (\Phimat\tsp\Phimat + \gamma \Imat)\inv [\Phimat\tsp \yv + \gamma(\xv+\bv)]}\,,
\end{eqnarray}
where $\Imat$ is an identity matrix with desired dimensions.
Since $\Phimat$ is a fat matrix, $(\Phimat\tsp\Phimat + \gamma \Imat)$ will be a large matrix and thus the matrix inversion formula is employed to simplify the calculation:
\begin{equation}
{(\Phimat\tsp\Phimat + \gamma \Imat)\inv = \gamma\inv \Imat - \gamma\inv \Phimat\tsp(\Imat + \Phimat\gamma\inv\Phimat\tsp)\inv \Phimat \gamma\inv\,.} \label{Eq:PhiTPhi_inv}
\end{equation} 
Plug~\eqref{Eq:PhiTPhi_inv} into~\eqref{Eq:ADMM_theta1}, we have
\begin{equation}\label{Eq:ADMM_theta2}
  \begin{aligned}
    \thetav =&{\,} \frac{[\Phimat\tsp \yv + \gamma(\xv+\bv)]}{\gamma}-\frac{\Phimat\tsp(\Imat + \Phimat\gamma\inv\Phimat\tsp)\inv \Phimat\Phimat\tsp \yv}{\gamma^{2}} \\
&{\,}\quad- \frac{\Phimat\tsp(\Imat + \Phimat\gamma\inv\Phimat\tsp)\inv \Phimat(\xv+\bv)}{\gamma}\,.
  \end{aligned}
\end{equation}
As mentioned earlier, $\Phimat\Phimat\tsp$ is a diagonal matrix in our imaging systems. 
Let
\begin{equation}
  {\Phimat\Phimat\tsp \stackrel{\rm def}{=}{\rm diag}\{\psi_1, \dots, \psi_n\}\,,} \label{Eq:phiphit}
\end{equation}
we have
\begin{align}
&  {(\Imat + \Phimat\gamma\inv\Phimat\tsp)\inv = {\rm diag}\left\{\frac{\gamma}{\gamma+ \psi_1}, \dots, \frac{\gamma}{\gamma + \psi_n}\right\}}\,,\\
&  {(\Imat + \Phimat\gamma\inv\Phimat\tsp)\inv\Phimat\Phimat\tsp = {\rm diag}\left\{\frac{\gamma \psi_1}{\gamma+ \psi_1}, \dots, \frac{\gamma\psi_n}{\gamma + \psi_n}\right\}}\,.
\end{align}
Let $\yv = [y_1,\dots,y_n]\tsp$ and $[\Phimat(\xv+\bv)]_i$ denote the $i^{th}$ element of the vector $\Phimat(\xv+\bv)$;~\eqref{Eq:ADMM_theta2} becomes 
\begin{align}
  {\thetav} =&{\,}   {\gamma\inv \Phimat\tsp \yv + (\xv+\bv )}\nonumber\\
&{\,}  {- \frac{\Phimat\tsp\big[\frac{y_1\psi_1 + \gamma[\Phimat(\xv+\bv)]_1}{\gamma+ \psi_1},\dots,\frac{y_n\psi_n + \gamma [\Phimat(\xv+\bv)]_n}{\gamma+ \psi_n}\big]\tsp}{\gamma}} \nonumber\\
=&{\,}   (\xv+\bv ) \nonumber\\
	&+ \Phimat\tsp\left[\frac{y_1 -[\Phimat(\xv+\bv)]_1}{\gamma+\psi_1},\dots,\frac{y_n -[\Phimat(\xv+\bv)]_n}{\gamma+\psi_n}\right]\tsp\,. \label{Eq:ADMM_theta3}
\end{align}
Note that $\{y_i -[\Phimat(\xv+\bv)]_i\}_{i=1}^n$ can be updated in one shot by $\yv - \Phimat(\xv+\bv)$, and $\{\psi_i\}_{i=1}^n$ is pre-calculated and stored with 
\begin{equation}
\psi_i = \sum_{k=1}^B c_{k,i}, \qquad \cv_k = {\rm {vec}}(\Cmat_k)
\end{equation}
with $\Cmat_k$ defined in~\eqref{Eq:YXC}. 
In this way, the last term in \eqref{Eq:ADMM_theta3} can be computed element-wise and
$\thetav$ can thus be updated very efficiently.

\textbf{Solve $\xv$:}
Let $\qv = \thetav-\bv$. Eq.~\eqref{Eq:x_sub} can be considered as the WNNM denoising problem (however, for videos rather than images)
\begin{equation}\label{Eq:ADMM_x1}
  {\hat{\xv} = \argmin_{\xv} \lambda \sum_i \|\Zmat_i\|_{w,*} + \frac{\gamma}{2} \|\qv - \xv\|_2^2\,.}
\end{equation}
Recall that $\Zmat_i$ is the $i^{th}$ patch group constructed from $\xv$. Let $\Rmat_i$ be the $i^{th}$ patch group constructed from $\qv$ corresponding to $\Zmat_i$.
The structure of \eqref{Eq:ADMM_x1} is very complicated and in order to achieve a tractable solution, a general assumption is made. Specifically, $\qv$ is treated as a noisy version of $\xv$, which is $\qv = \xv + \ev$, where $\ev$ denotes the zero-mean white Gaussian noise and each element $e_i$ follows an independent zero mean Gaussian
distribution with variance $\sigma^2_n$, \ie, $e_i \sim {\cal N}(0, \sigma^2_n)$. It is worth noting that we can recover $\xv$ after reconstructing $\{\Zmat_i\}_{i=1}^N$, and there are in total $N$ patch groups. 
Invoking the law of large numbers in probability theory, we have the following equation with a very large probability (limited to 1) at each iteration~\cite{Zha17ICME}, \ie,
$\forall \varepsilon>0$, 
\begin{equation}
\lim_{\substack{nB\rightarrow\infty\\dMN\rightarrow\infty}} p{\Big(\frac{1}{nB}\|\qv - \xv\|_2^2 - \frac{1}{dMN}\sum_{i=1}^N\|\Rmat_i -\Zmat_i \|_\mathrm{F}^2<\varepsilon\Big)}=1.
\end{equation}
The proof can be found in~\cite{Zhang14TIP}.
As both $n$ and $N$ are large (\ie, $>60000$) in our case and the overlapping patches are averaged, we can thus  translate \eqref{Eq:ADMM_x1} to the following problem
\begin{equation}\label{Eq:ADMM_x2}
  {\hat{\xv} = \argmin_{\xv} \lambda \sum_i \|\Zmat_i\|_{w,*} + \frac{\gamma}{2} \sum_i\|\Rmat_i -\Zmat_i \|_\mathrm{F}^2\,.}
\end{equation}
Note that there is a scale difference between $\gamma$ in (\ref{Eq:ADMM_x2}) and (\ref{Eq:ADMM_x1}). As mentioned above, $\xv$ can be recovered after reconstructing $\{\Zmat_i\}_{i=1}^N$ and each $\Zmat_i$ can be independently solved by
\begin{equation} \label{Eq:ADMM_Zi}
  {\hat{\Zmat_i} = \argmin_{\Zmat_i} \frac{1}{2} \|\Rmat_i -\Zmat_i \|_\mathrm{F}^2 + \frac{\lambda}{\gamma} \|\Zmat_i\|_{w,*}\,.}
\end{equation} 
By considering $\Rmat_i = \Zmat_i + \Emat_i$ and $\Emat_i \sim {\cal N}({\bf 0}, \sigma_n^2 \Imat)$, \eqref{Eq:ADMM_Zi} can be solved in closed-form by the generalized soft-thresholding algorithm~\cite{Zuo13ICCV_GIST,Gu17IJCV} detailed below, where we have used $\sigma_n^2 = \lambda/\gamma$.

Considering the singular value decomposition (SVD) of { $\Rmat_i = \Umat \Sigmamat \Vmat\tsp$ and the weight vector $0\le w_1  \le \dots \le w_{min\{d,m\}}$, $\hat{\Zmat}_i$ can be solved by
\begin{eqnarray}
{\hat{\Zmat}}_i &=& \Umat\mathcal{D}_{\wv}(\Sigmamat)\Vmat\tsp\,, \label{Eq:Zest_1}\\
\mathcal{D}_{\wv}(\boldsymbol\Sigma)_{j,j} &=& \max \{\Sigmamat_{j,j} - w_j,0\}\,.  \label{Eq:Zest_2}
\end{eqnarray}
} 
The following problem is to determine the weight vector $\wv$.
For natural images/videos, we have the general
prior knowledge that the larger singular values of $\Zmat_i$ are
more important than the smaller ones since they represent
the energy of the major components of $\Zmat_i$. In the application
of denoising, the larger the singular values, the less
they should be shrunk, \ie,
\begin{equation}
\label{equ:weight}
  {w_j = \frac{c\sqrt{M}}{\sigma_j(\Zmat_i) + \epsilon}}\,,
\end{equation}
where $c>0$ is a constant, $M$ is the number of patches in $\Zmat_i$ and $\epsilon>0$ is a tiny positive constant. 
Since $\xv$ is not available, $\sigma_j(\Zmat_i)$ is not available either. We estimate it by 
\begin{equation}
  {\hat{\sigma}_j(\Zmat_i) = \sqrt{{\rm max}\{\sigma^2_j(\Rmat_i) - M\sigma^2_n,0\}}}\,, \label{Eq:ADMM_sigma}
\end{equation}
where $\sigma_n^2$ is updated in each iteration. There are some heuristic approaches on how to determine $\sigma_n^2$ based on current estimations~\cite{Gu14CVPR}. In principle, with the increasing of the iteration number, the measurement error is decreasing and $\sigma_n^2$ is getting smaller.
In our SCI problem, we have found that progressively decreasing the noise level by starting with a large value performs well in our experiments (details in Sec.~\ref{ssub:para_set}). This is consistent with the analysis in~\cite{Gu14CVPR}. 

The complete algorithm, \ie, DeSCI, is summarized in Algorithm~\ref{algo:ADMM-WNNM}. 
\begin{center}
  \begin{algorithm}[!t]
    \caption{The DeSCI Algorithm for SCI reconstruction}
    \begin{algorithmic}[1]
      \REQUIRE$\Phimat$, $\yv$
      \STATE Initialize $\lambda>0$, $\gamma>0$, $c>0$, $\epsilon = 10^{-16}$, $\sigma_n$, $\xv =\Phimat\tsp\yv$, and $\bv = {\bf 0}$.
      \STATE Pre-calculate $\Phimat\Phimat\tsp$ and save the diagonal elements $\{\psi_i\}_{i=1}^n$.
      \FOR{$t=0$ \TO Max-Iter }
      \STATE Update $\thetav$ by Eq.~\eqref{Eq:ADMM_theta3}.
      \STATE Update $\qv = \thetav - \bv$.
      \FOR {Each patch $\rv_i$ in $\qv$}
      \STATE
      Find similar patches to construct patch group $\Rmat_i$.
      \STATE
      Perform SVD to $\Rmat_i$.
      \STATE
      Estimate $\Zmat_i$ using Eqs.~\eqref{Eq:Zest_1}-\eqref{Eq:ADMM_sigma}.
      \ENDFOR
      \STATE Aggregate $\{\Zmat_i\}_{i=1}^N$ to form $\xv$.
      \STATE
       Update $\sigma_n$.
       \STATE Update $\bv$ by Eq.~\eqref{Eq:ADMM_b1}.
      \ENDFOR
      \STATE Output: the recovered video (hyperspectral images) $\xv$.
    \end{algorithmic}
    \label{algo:ADMM-WNNM}
  \end{algorithm}
\end{center}

\textbf{Relation to generalized alternating projection. \label{Sec:GAP}}
The generalized alternating projection (GAP) algorithm~\cite{Liao14GAP} has been investigated extensively for video CS~\cite{Patrick13OE,Yuan14CVPR,Yuan16ICIP_GAP} with different priors. 
The difference between ADMM developed above and GAP is the constraint on the measurement. In ADMM, we aim to minimize $\frac{1}{2}\|\yv-\Phimat\xv\|_2^2$, while GAP imposes $\yv = \Phimat\xv$.  
Specifically, by introducing a constant $C$, GAP reformulates the problem in~\eqref{eq:Admm_WNNM_1} as
\begin{eqnarray}\label{Eq:gap1}
  {\min_{\xv}\, C \quad\st~ \sum_i\|\Zmat_i\|_{w,*}  \le C ~~{\rm and}~~ \Phimat\xv = \yv}\,.
\end{eqnarray} 
This is solved by 
a series of alternating projection problem:
\begin{equation}\label{Eq:gap2}
  \begin{aligned}
      &{\left(\xv^{(t)}, \thetav^{(t)}\right)} ={~}   {\argmin_{(\xv,\thetav)} \frac{1}{2}\|\xv-\thetav\|^2_2}\,,\\
    &\qquad\quad{~}\;\st\;\;    {\sum_i\|\Rmat_i\|_{w,*}}\le   {C^{(t)}~~\text{and}~~ \Phimat\xv = \yv}\,,
  \end{aligned}
\end{equation}
where $C^{(t)}$ is a constant that changes in each iteration; $\Rmat_i$ is constructed from $\thetav$, and $\Zmat_i$ is constructed from $\xv$, respectively.

Eq.~\eqref{Eq:gap2} is equivalent to
\begin{equation}\label{eq:gapregu}
  \begin{aligned}
      {\left(\xv^{(t)}, \thetav^{(t)}\right)} =&{~}  {\argmin_{(\xv,\thetav)} \frac{1}{2}\|\xv-\thetav\|_2^2 + \beta \sum_i\|\Rmat_i\|_{w,*}}\,, \\
    &{~}\quad\;\st\;\;  \Phimat\xv = \yv\,,
  \end{aligned}
\end{equation}
where $\beta$ is a parameter to balance these two terms.
%
Eq.~\eqref{eq:gapregu} is solved by alternating updating $\thetav$ and $\xv$.
Given $\xv$, $\thetav$ is solved by a WNNM denoising algorithm as derived before.

Given $\thetav$, the update of $\xv$ is simply an Euclidean projection of $\thetav$ on the linear manifold ${\cal M}: \yv =\Phimat\xv$; this is solved by
\begin{align} %
\xv^{(t)}&= \thetav^{(t-1)} + \Phimat\tsp(\Phimat\Phimat\tsp)^{-1}(\yv - \Phimat\thetav^{(t-1)})\,,  \label{eq:xvt} \\
&\stackrel{\eqref{Eq:phiphit}}{=}  \thetav^{(t-1)} + \Phimat\tsp \left[\frac{y_1^{(t-1)} -[\Phimat\thetav^{(t-1)}]_1}{\psi_1}, \right.\nonumber\\
	&\qquad \qquad \qquad\left.\dots,\frac{y_n^{(t-1)} -[\Phimat\thetav^{(t-1)}]_n}{\psi_n}\right]\tsp\,. \label{eq:gap_xt}
\end{align}
Comparing with the updated equation of $\thetav$ in \eqref{Eq:ADMM_theta3}, by interchange $\xv$ and $\thetav$, we can found \eqref{Eq:ADMM_theta3} and \eqref{eq:gap_xt} are equivalent by setting $\bv = {\bf 0}$ and $\gamma=0$ in \eqref{Eq:ADMM_theta3}. This is expected as if we set $\gamma=0$ in \eqref{Eq:theta_sub}, we impose that $\thetav$ is on the linear manifold ${\cal M}: \yv =\Phimat\xv$, which is exactly the constraint used in GAP. This works well in the noiseless and low noise cases. However, when the noise level is high, ADMM will outperform GAP as the strong constraint $\yv = \Phimat\xv$ will bias the results. 

An accelerated GAP (GAP-acc) is also proposed by Liao {\em et al.}~\cite{Liao14GAP}. The linear manifold can also be adaptively adjusted and so \eqref{eq:xvt} can be modified as
\begin{eqnarray} \label{Eq:acc_gap}
\xv^{(t)}&=& \thetav^{(t-1)} + \Phimat\tsp(\Phimat\Phimat\tsp)^{-1}(\yv^{(t-1)} - \Phimat\thetav^{(t-1)})\,,\\
\yv^{(t)}&=& \yv^{(t-1)} + (\yv - \Phimat \thetav^{(t-1)})\,. \quad \forall t\ge 1\,.
\end{eqnarray}
This accelerated GAP speeds up the convergence in the noiseless case in our experiments. 
We have also tried the approximate message passing (AMP) algorithm~\cite{Donoho09AMP} used in~\cite{Mertzler14Denoising} to update $\xv$. Unfortunately, AMP cannot lead to good results due to the special structure of $\Phimat$; this may be because the convergence of AMP heavily depends on the Gaussianity of the sensing matrix in CS.

\section{Simulation results \label{Sec:Sim_Results}}
In this section, we validate the proposed DeSCI algorithm on simulation datasets, including the videos and hyperspectral images and compare it with other state-of-the-art algorithms for SCI reconstruction.  

\subsection{Video snapshot compressive imaging \label{Sec:sim_video}}
We first conduct simulations to the video SCI problem. 
The coding strategy employed in~\cite{Patrick13OE}, specifically the shifting binary mask, is used in our simulation.
Two datasets, namely \texttt{Kobe} and \texttt{Traffic}, used in~\cite{Yang14GMM} are employed in our simulation. We also used another two datasets, \texttt{Runner}~\cite{Runner} and \texttt{Drop}~\cite{Drop} to show the wide applicability.  
The video frame is of size $256\times 256$ (pixels) and eight ($B=8$) consequent frames are modulated by shifting masks and then collapsed to a single measurement. 
We compare our proposed DeSCI algorithm with other leading algorithms, including {GMM-TP}~\cite{Yang14GMM}, {MMLE-GMM}, {MMLE-MFA}~\cite{Yang14GMMonline} and GAP-TV~\cite{Yuan16ICIP_GAP}. The performance of GAP-wavelet proposed in~\cite{Yuan14CVPR} is similar to GAP-TV as demonstrated in~\cite{Yuan16ICIP_GAP} and thus omitted here.
Due to the special requirements of the mask in deep learning based methods~\cite{Iliadis18DSPvideoCS}, we do not compare with them. However, as mentioned earlier, we can notice that limited improvement (around 2 dB) has been obtained using these deep learning techniques in~\cite{Iliadis18DSPvideoCS} compared with GMM. By contrast, our algorithm has improved the results significantly ($>4$ dB) over GMM.
All algorithms are implemented in MATLAB and the GMM and GAP-TV codes are downloaded from the authors' websites.
The MATLAB code of the proposed DeSCI algorithm is available at \textcolor{blue}{https://github.com/liuyang12/DeSCI.}

\subsubsection{Parameter setting} 
\label{ssub:para_set}
In our DeSCI algorithm, for all noise levels, the parameter $c$ and iteration number Max-Iter are fixed to $2.8$ and $60$, respectively.  
The image patch size is set to $6\times6,7\times7,8\times8$ and $9\times9$ for $\sigma_n\le20,20<\sigma_n\le40,40<\sigma_n\le60$ and $\sigma_n>60$, respectively; the searching window size is set to $30$, and the number of adjacent frames for denoising is set to $8$. 
Each patch group has $70, 90, 120$ and $140$ patches for $\sigma_n\le20,20<\sigma_n\le40,40<\sigma_n\le60$ and $\sigma_n>60$, respectively. 
The parameter $\gamma$ is set based on the signal-to-noise ratio (SNR) of the SCI measurements. Specifically, $\gamma$ is set to $\{0.24, 1.2, 6, 30, 150\}$ for the measurement SNR of \{40dB, 30dB, 20dB, 10dB, 0dB\}. For noiseless measurements, $\gamma$ is set to $0$; as derived in Sec.~\ref{Sec:GAP}, in this case, ADMM is equivalent to GAP.
Our algorithm is terminated by sufficiently small difference between successive updates of $\xv$ or reaching the Max-Iter.
When the SNR of the SCI measurement is larger than 30 dB, GAP-acc is used to update $\xv$. Both PSNR and structural similarity (SSIM)~\cite{Wang04imagequality} are employed to evaluate the quality of reconstructed videos.
\begin{figure}[!t]
  \centering
  {\includegraphics[width=\columnwidth]{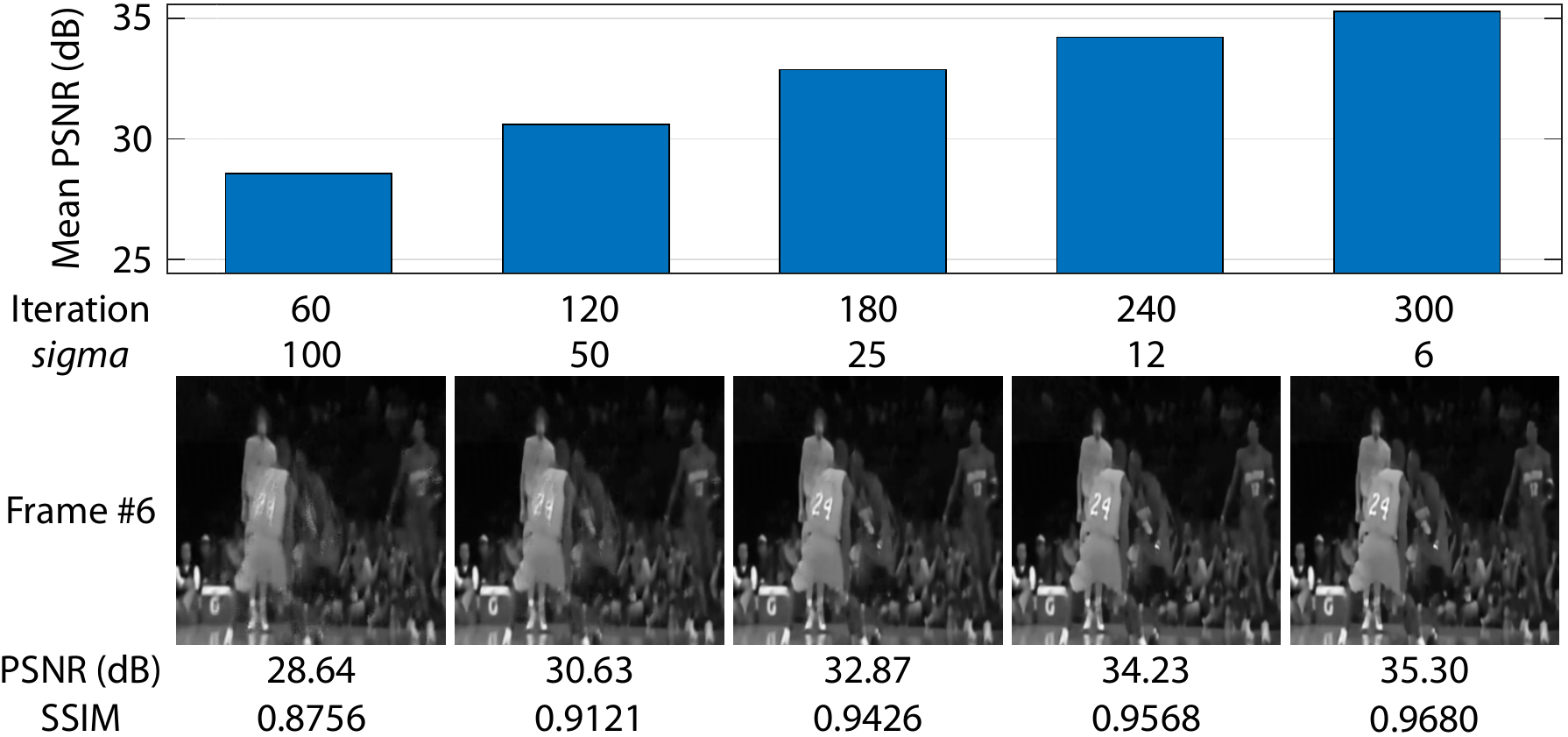}}
  \caption{ The PSNR of reconstructed video frames by setting different $\sigma_n$ at different iterations of DeSCI. It can be seen that the $\sigma_n$ is starting from a large value (100) and then decreasing by half for every 60 iterations. Both PSNR and SSIM are gradually increasing with the iteration number. }
  \label{fig:PSNR_kobe_sigma}
\end{figure}

As mentioned earlier, the noise estimation in each iteration of our DeSCI algorithm is important to the performance.
We found that firstly setting $\sigma_n$ to a large value and then progressively decreasing it leads to good results. One example of the \texttt{Kobe} data used in~\cite{Yang14GMM} is shown in Fig.~\ref{fig:PSNR_kobe_sigma}. We have found that this sequential setting can always lead to good results in our experiments.

\subsubsection{Video compressive imaging results} 
\label{ssub:video_compressive_imaging_results}

\begin{table*}[!t]
  \caption{The average results of PSNR in dB (left entry in each cell) and SSIM (right entry in each cell) by different algorithms on 4 datasets.}
  \centering
  {
    \begin{tabular}{|c|c|c|c|c|c|}
      \hline
      Algorithm& \texttt{Kobe} & \texttt{Traffic} & \texttt{Runner} & \texttt{Drop} & Average\\
      \hline
      GMM-TP & 24.47, 0.5246 & 25.08, 0.7652 & 29.75, 0.6995 & 34.76, 0.6356 & 28.52, 0.6562\\
      \hline
      MMLE-GMM & 27.33, 0.6962 & 25.68, 0.7798 & 33.68, 0.8224 & 39.86, 0.7369 & 31.64, 0.7588\\
      \hline
      MMLE-MFA & 24.63, 0.5291 & 22.66, 0.6232 & 30.83, 0.7331 & 35.66, 0.6690 & 28.45, 0.6386\\
      \hline
      GAP-TV & 26.45, 0.8448 & 20.89, 0.7148 & 28.81, 0.9092 & 34.74, 0.9704& 27.72, 0.8598\\
      \hline
      DeSCI & {\bf 33.25}, {\bf 0.9518} & {\bf 28.72}, {\bf 0.9251} & {\bf 38.76}, {\bf 0.9693} & {\bf 43.22}, {\bf 0.9925} & {\bf 35.99}, {\bf 0.9597}
      \\
      \hline
  \end{tabular}}
  \label{Tab:results_4video}
\end{table*}

We hereby demonstrate that the proposed DeSCI algorithm performs much better than current state-of-the-art algorithms. Comparison of DeSCI, GMM-TP, MMLE-GMM, MMLE-MFA and GAP-TV on four datasets is shown in Table~\ref{Tab:results_4video}. 
It can be seen clearly that the proposed DeSCI outperforms other algorithms on every dataset. Specifically, the average gains of DeSCI over GMM-TP, MMLE-GMM, MMLE-MFA and GAP-TV are as much as \{7.47, 4.35, 7.54, 8.27\}dB on PSNR and \{0.3035, 0.2009, 0.3211, 0.0999\} on SSIM. 
\begin{figure*}[htbp!]
  \centering
  {\includegraphics[width=2\columnwidth]{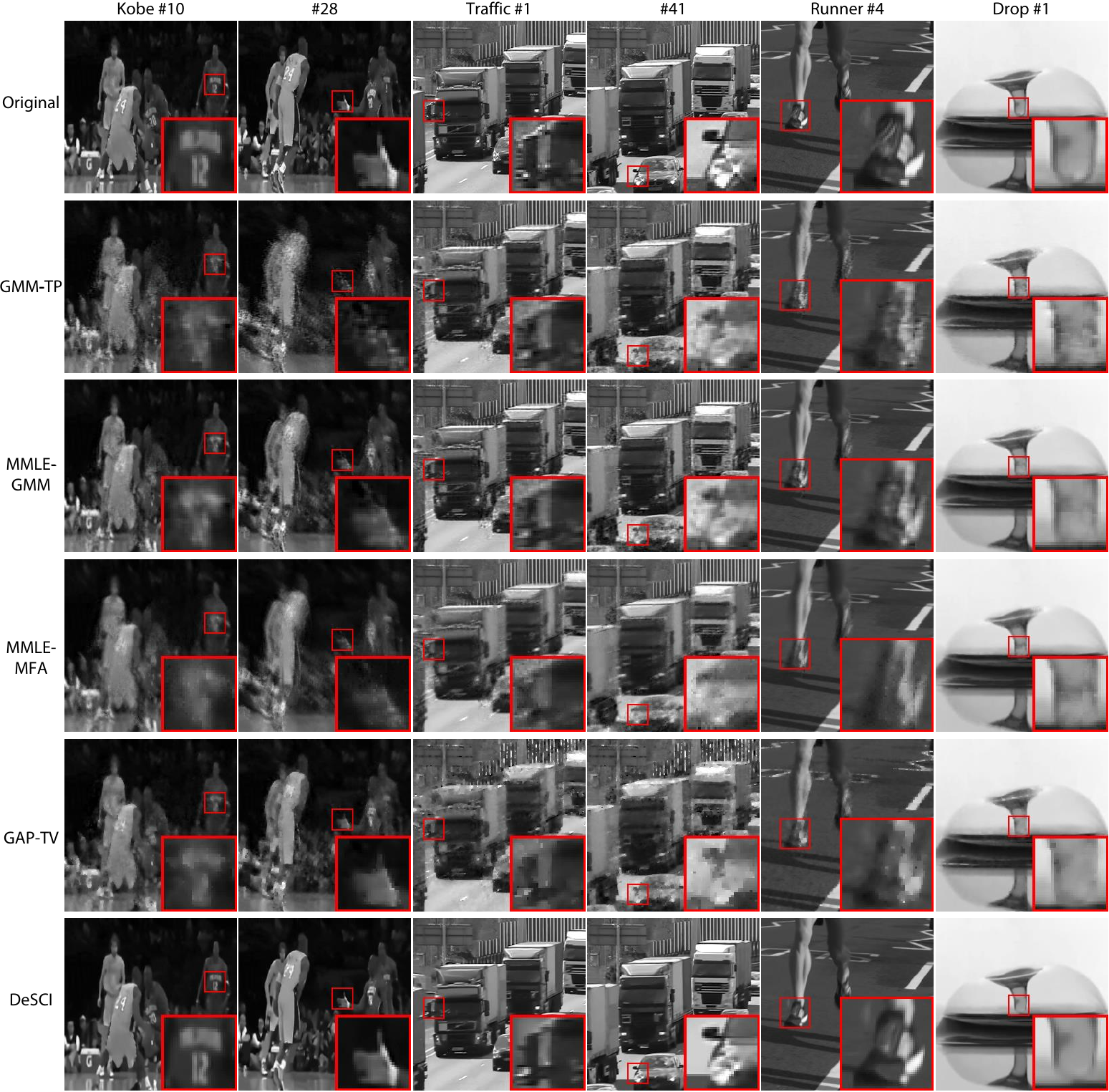}}
  \caption{Reconstruction frames of DeSCI and other algorithms (GMM-TP, MMLE-GMM, MMLE-MFA, and GAP-TV).
  }
  \label{fig:gmm_gaptv_desci_frames}
\end{figure*}
In average, MMLE-GMM is the runner-up on PSNR while GAP-TV is the runner-up on SSIM. This may due to the fact that the TV prior is imposed globally and thus reserves more structural information of the video (thus leading to a higher SSIM).
Exemplar reconstructed frames are shown in Fig.~\ref{fig:gmm_gaptv_desci_frames}.
It can be observed that GMM based algorithms suffer from blurring artifacts and GAP-TV suffers from blocky artifacts. By contrast, our proposed DeSCI algorithm can not only provide the fine details but also the large-scale sharp edges and clear motions.  
The reconstructed videos are shown in the supplementary material (SM). 

\begin{figure}[!htbp]
  \centering
  {\includegraphics[width=\columnwidth]{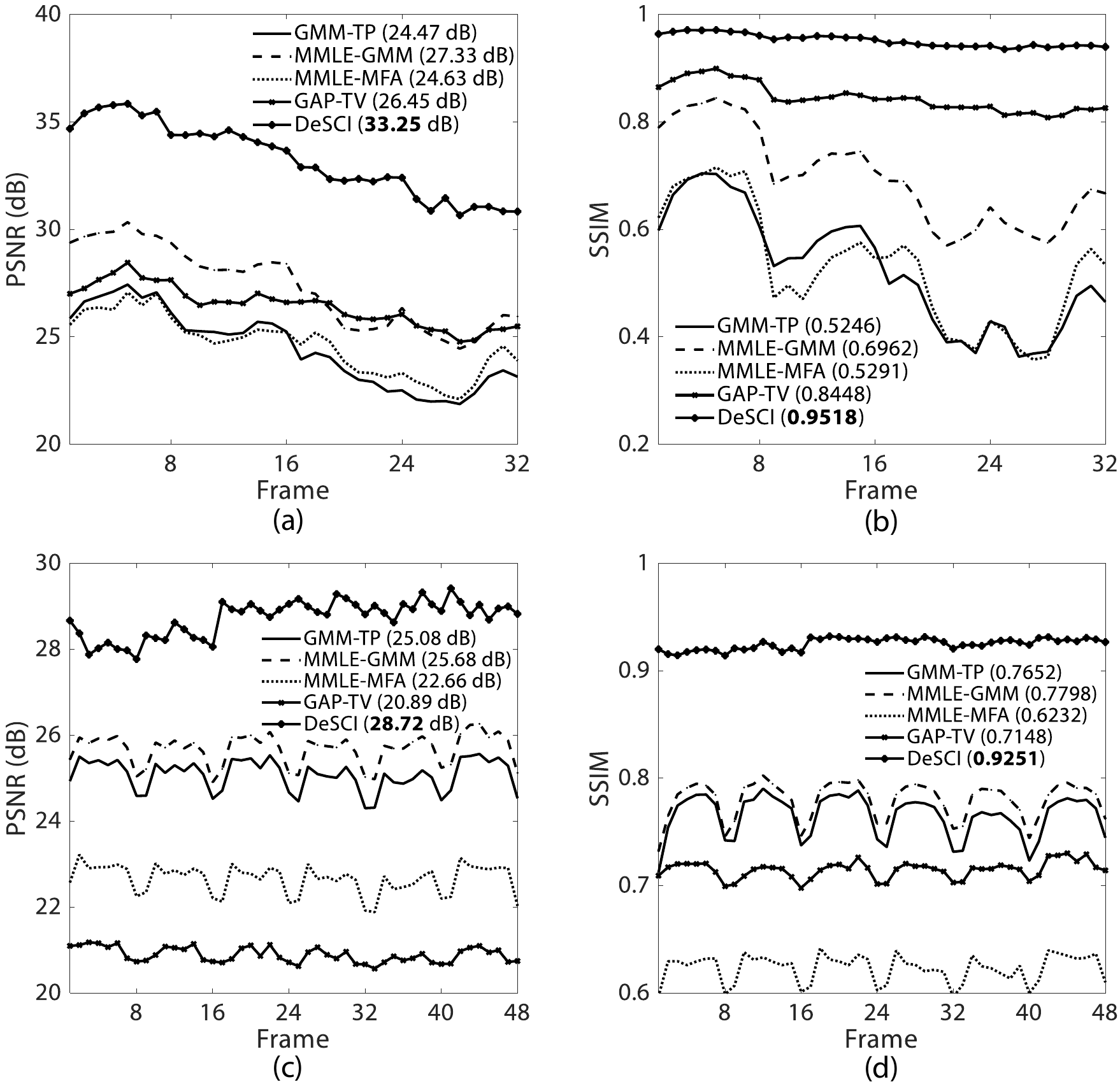}}
  \caption{ Frame-wise PSNR (a,c) and SSIM (b,d) of DeSCI and other algorithms for the \texttt{Kobe} (a-b) and \texttt{Traffic} (c-d) datasets. }
  \label{fig:gmm_gaptv_desci_kobe}
\end{figure}

One phenomenon for SCI reconstruction is the image quality (PSNR and SSIM) dropping down of the first and last reconstruction frames for each measurement. 
One of the reasons is that previous reconstruction algorithms do not consider the non-local self similarity in video frames. 
In Fig.~\ref{fig:gmm_gaptv_desci_kobe}, we plot the frame-wise PSNR and SSIM of DeSCI as well as other algorithms for the \texttt{Kobe} and \texttt{Traffic} datasets. It can be seen that the proposed DeSCI smooths this quality dropping down phenomenon out. 
The self-similarity exists in every frame of the video and thus helps the reconstruction of these frames. 
We can also notice that the PSNR of reconstructed video frames drops in the last 16 frames of the \texttt{Kobe} data (Fig.~\ref{fig:gmm_gaptv_desci_kobe}(a)); this is because complicated motions exist in these frames, \ie, the slam dunk. 
Though the PSNRs of all algorithms drop, the SSIM of our proposed DeSCI is much smoother (Fig.~\ref{fig:gmm_gaptv_desci_kobe}(b)) than that of other algorithms. 

\subsubsection{Robustness to noise} 
\label{ssub:robustness_to_noise}
\begin{figure}[!t]
  \centering
  {\includegraphics[width=\columnwidth]{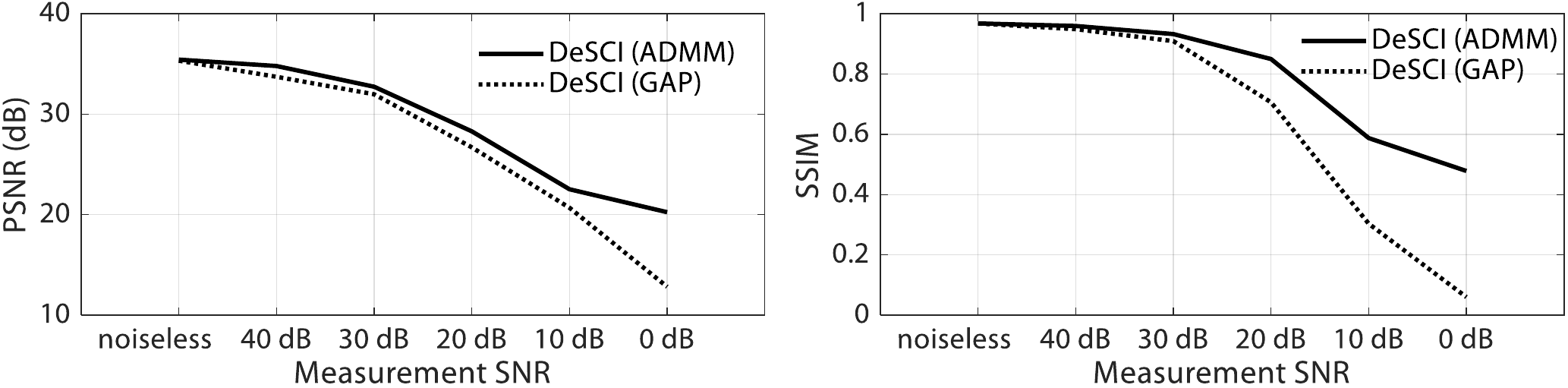}}
  \caption{ Comparison of ADMM and GAP for DeSCI with noisy measurements.  A single measurement of the \texttt{Kobe} dataset is used. }
  \label{fig:admm_vs_gap_kobe}
\end{figure}

As mentioned in Sec.~\ref{Sec:ADMM}, DeSCI using GAP as projection can lead to fast results under noiseless case, which has been verified in previous results.
However, GAP imposes that the solution of $\xv$ should be on the manifold ${\cal M}: \yv = \Phimat \xv$, which is too strong in the noisy case.
By contrast, ADMM minimizes the measurement error, \ie, $\frac{1}{2}\|\yv-\Phimat\xv\|_2^2$, and is thus robust to noise~\cite{Yin08bregman,Boyd11ADMM}.
To verify this, we perform the experiments on the \texttt{Kobe} dataset by adding different levels of white Gaussian noise.
The results are summarized in Fig.~\ref{fig:admm_vs_gap_kobe}, where we can see that in the noiseless case, ADMM and GAP performs the same, which is consistent with our theoretical analysis. When the measurement SNR is getting smaller, ADMM outperforms GAP in both PSNR and SSIM.
Therefore, DeSCI using ADMM as projection is recommended in realistic systems with noise.

\subsubsection{DeSCI with other denoising algorithms} 
\label{ssub:desci_with_other_denoising_algorithms}
\begin{figure}[!t]
  \centering
  {\includegraphics[width=\columnwidth]{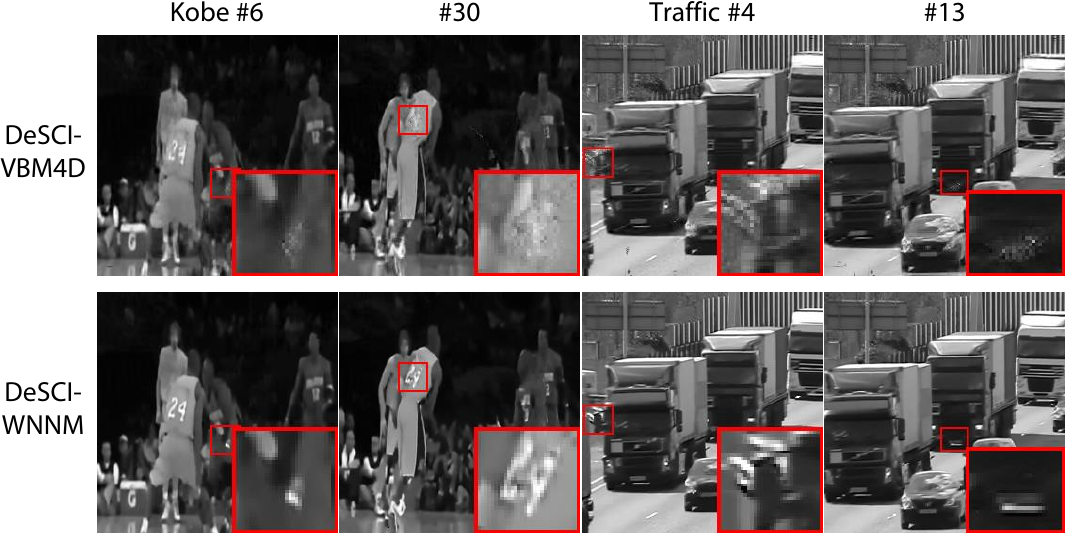}}
  \caption{ Reconstruction frames of DeSCI using VBM4D and WNNM for video denoising. We can see that DeSCI-VBM4D still suffers from some undesired artifacts while DeSCI-WNNM can provide fine details as well as large-scale sharp edges. }
  \label{fig:wnnm_vs_vbm4d_frames}
\end{figure}

Our proposed DeSCI is an iterative algorithm and in each iteration, we first perform projection via ADMM or GAP and then perform denoising via the low-rank minimization method, namely, WNNM. 
A natural question is how will the performance change by replacing the derived WNNM with other state-of-the-art video denoising algorithms. 
We have already seen that the DeSCI-WNNM performs much better than TV based denoising algorithms. The VBM4D~\cite{Maggioni2012VideoDD} algorithm, which is the representative state-of-the-art video denoising algorithm, is incorporated into our proposed framework, dubbed DeSCI-VBM4D. 
We run DeSCI-VBMD on the \texttt{Kobe} and \texttt{Traffic} datasets and obtained the PSNR of \{30.60, 26.60\}dB and SSIM of \{0.9260, 0.8958\}, respectively.
More specifically,  DeSCI-WNNM outperforms DeSCI-VBM4D more than 2 dB on PSNR and more than 0.025 on SSIM.
Exemplar reconstruction frames are shown in Fig.~\ref{fig:wnnm_vs_vbm4d_frames}, where we can see DeSCI-VBM4D still suffers from some undesired artifacts while the DeSCI-WNNM can provide small-scale fine details as well as large-scale sharp edges.
This clearly verifies the superiority of the proposed reconstruction framework along with the WNNM denoising on SCI reconstruction.  
     
We have also tried to integrate WNNM and VBM4D with other priors for denoising, such as TV priors. We observe that the TV prior can marginally help DeSCI-VBM4D (less than 0.2 dB) but almost cannot help DeSCI-WNNM. 
This again verifies that our proposed DeSCI-WNNM has investigated both the global prior and the local prior for videos, through the nonlocal self-similarity and low-rank estimation. 

\subsubsection{Computational complexity} 
\label{ssub:computational_complexity}
As mentioned in Sec.~\ref{Sec:ADMM}, the projection step can be updated very efficiently, and the time consuming step is the weighted nuclear norm minimization for video denoising, in which both the block matching and low-rank estimation via SVD require a high computational workload. While the block matching can be performed once per (about) 20 iterations, the SVD needs to be performed in every iteration for each patch group. 
Specifically, a single round of block matching and patch estimation via SVD takes 90 seconds and 11 seconds, respectively. 
Recent advances of deep learning have shown great potentials in various image processing tasks, and we envision the block matching might be able to speed up via Generative Adversarial Networks (GAN)~\cite{NIPS2014_5423}.  In this way, the time of block matching can be reduced significantly.  
Regarding the low-rank estimation via SVD, some truncated approaches~\cite{Halko11SIAM} can be employed to speed-up the operation. 
Via using all these advanced tools, we believe each iteration of our algorithm can be performed within 10 seconds, and DeSCI can provide good results in a few minutes.



\begin{table}[htbp!]
	\caption{The average results of PSNR (left entry in each cell) and SSIM (right entry in each cell) by GAP-TV and DeSCI on simulated \texttt{bird} and \texttt{toy} hyperspectral datasets.}
	\centering
	{
		\begin{tabular}{|c|c|c|c|c|c|}
			\hline
			Algorithm& \texttt{bird} & \texttt{toy} \\
			\hline
			GAP-TV & 30.36 dB, 0.9251 & 24.66 dB, 0.8608 \\
			\hline
			DeSCI & {\bf 32.40} dB, {\bf 0.9452} & {\bf 25.91} dB, {\bf 0.9094} 
			\\
			\hline
	\end{tabular}}
	\label{Tab:results_sim_hsi}
\end{table}

\begin{figure}
	\centering
	{\includegraphics[width=1\columnwidth]{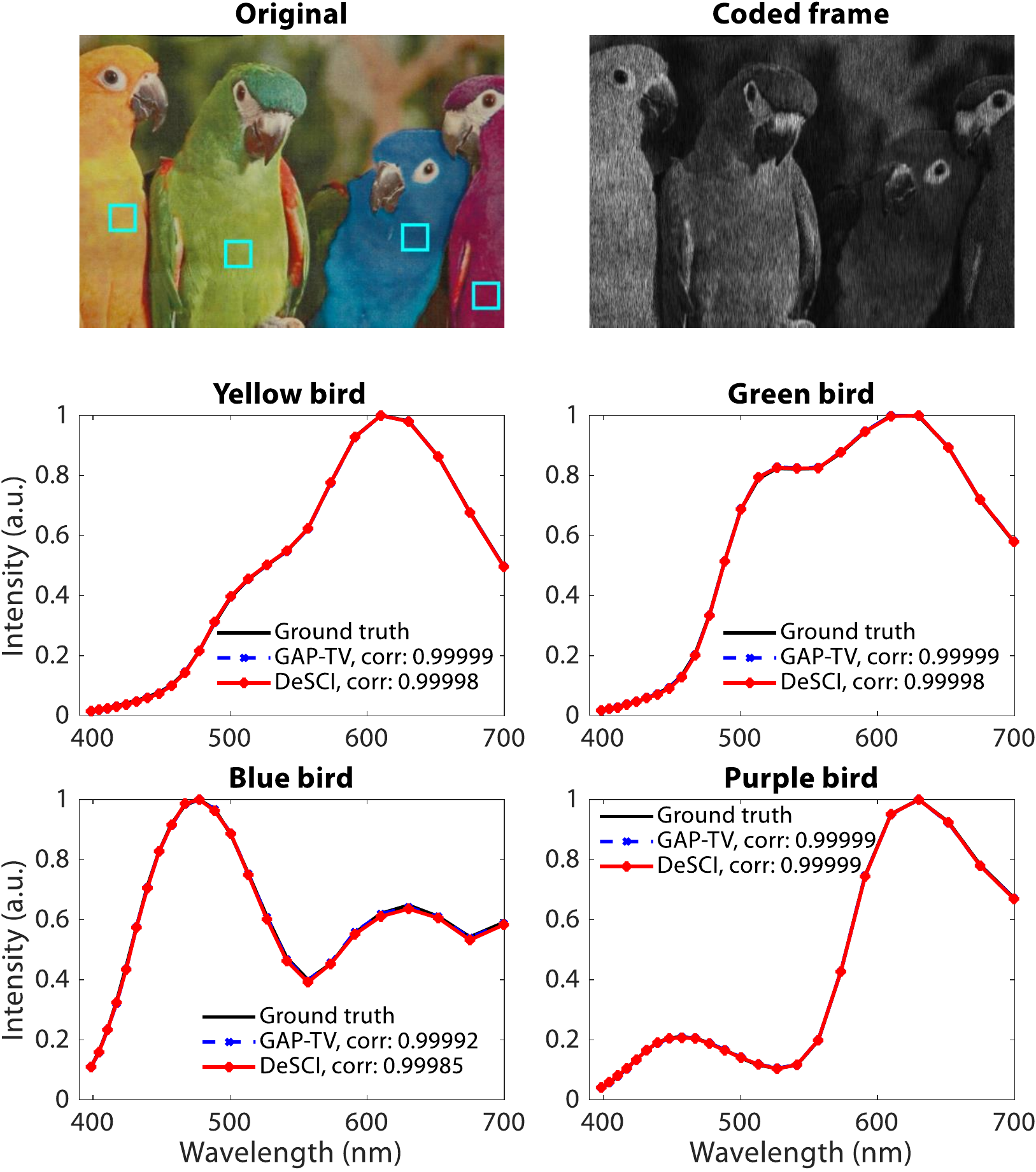}}
	\caption{Reconstructed spectra of simulated \texttt{bird} hyperspectral data. A snapshot measurement encoding 24 spectral bands is shown on the top-right. The original RGB image of the scene is shown on the top-left with a size of $1021\times703$ pixels. The spectra of four birds are shown on the middle and bottom rows. The correlation of the reconstructed spectra and the ground truth is shown in the legends.}
	\label{sfig:sim_hsi_bird_spectra}
\end{figure}
\begin{figure}
	\centering
	{\includegraphics[width=1\columnwidth]{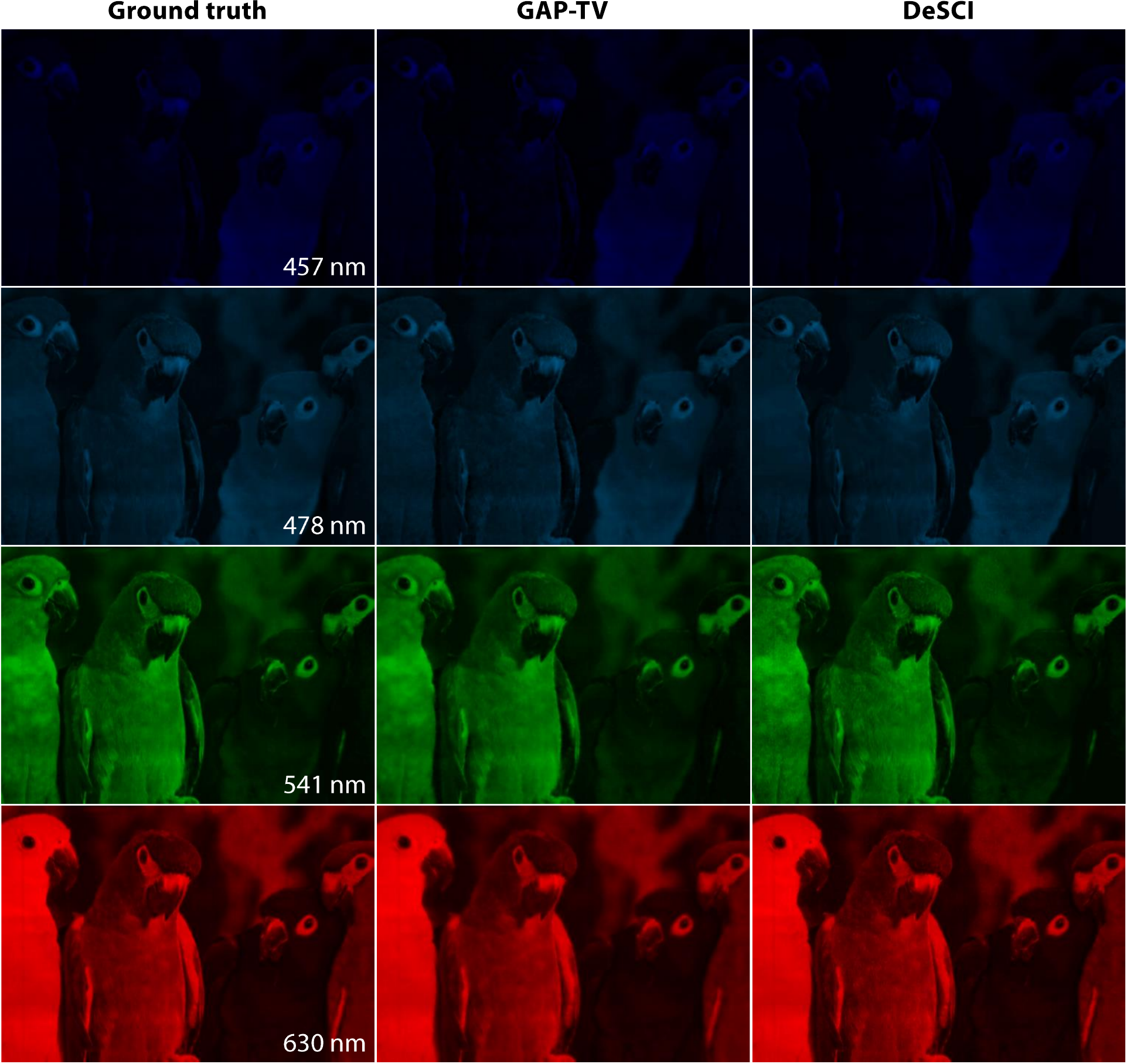}}
	\caption{Reconstructed frames of simulated \texttt{bird} hyperspectral data.}
	\label{sfig:sim_hsi_bird_frames}
\end{figure}
\subsection{Snapshot compressive hyperspectral image  \label{Sec:sim_hyper}}
We further demonstrate that DeSCI can be used for snapshot-hyperspectral compressive imaging systems, such as CASSI~\cite{Gehm07}. Both simulated data with shifting masks and data from CASSI systems~\cite{Gehm07} are presented. We first show the simulation results in this section and real data results are demonstrated in Sec.~\ref{Sec:real_hyper}. 
We generate the simulated hyperspectral data by summing the spectral frames from the hyperspectral dataset\footnote{The \texttt{bird} spectral data is from~\cite{Yuan15JSTSP}. The \texttt{toy} spectral data is from CAVE multispectral image database (http://www1.cs.columbia.edu/CAVE/databases/multispectral/).} and shifting random masks using Eq.~\eqref{Eq:YXC}. 

{\bf{ \texttt{Bird} simulated hyperspectral data.}} The \texttt{bird} hyperspectral images consists of 24 spectral bands with each of size $1021\times703$ pixels. The results of the reconstructed spectra of simulated \texttt{bird} hyperspectral data and exemplar frames are shown in Fig.~\ref{sfig:sim_hsi_bird_spectra} and Fig.~\ref{sfig:sim_hsi_bird_frames}, respectively. The averaged results of PSNR and SSIM are shown in Table~\ref{Tab:results_sim_hsi}. It can be seen clearly that both GAP-TV and DeSCI can recover the spectra of the four birds reliably with correlation over 0.9999, as shown in the legends of Fig.~\ref{sfig:sim_hsi_bird_spectra}. DeSCI provides more details in reconstruction (see Fig.~\ref{sfig:sim_hsi_bird_frames}, where the feather of the orange bird is less smoothed out than GAP-TV) and higher quantitative indexes (PSNR and SSIM in Table~\ref{Tab:results_sim_hsi}).

\begin{figure}
	\centering
	{\includegraphics[width=1\columnwidth]{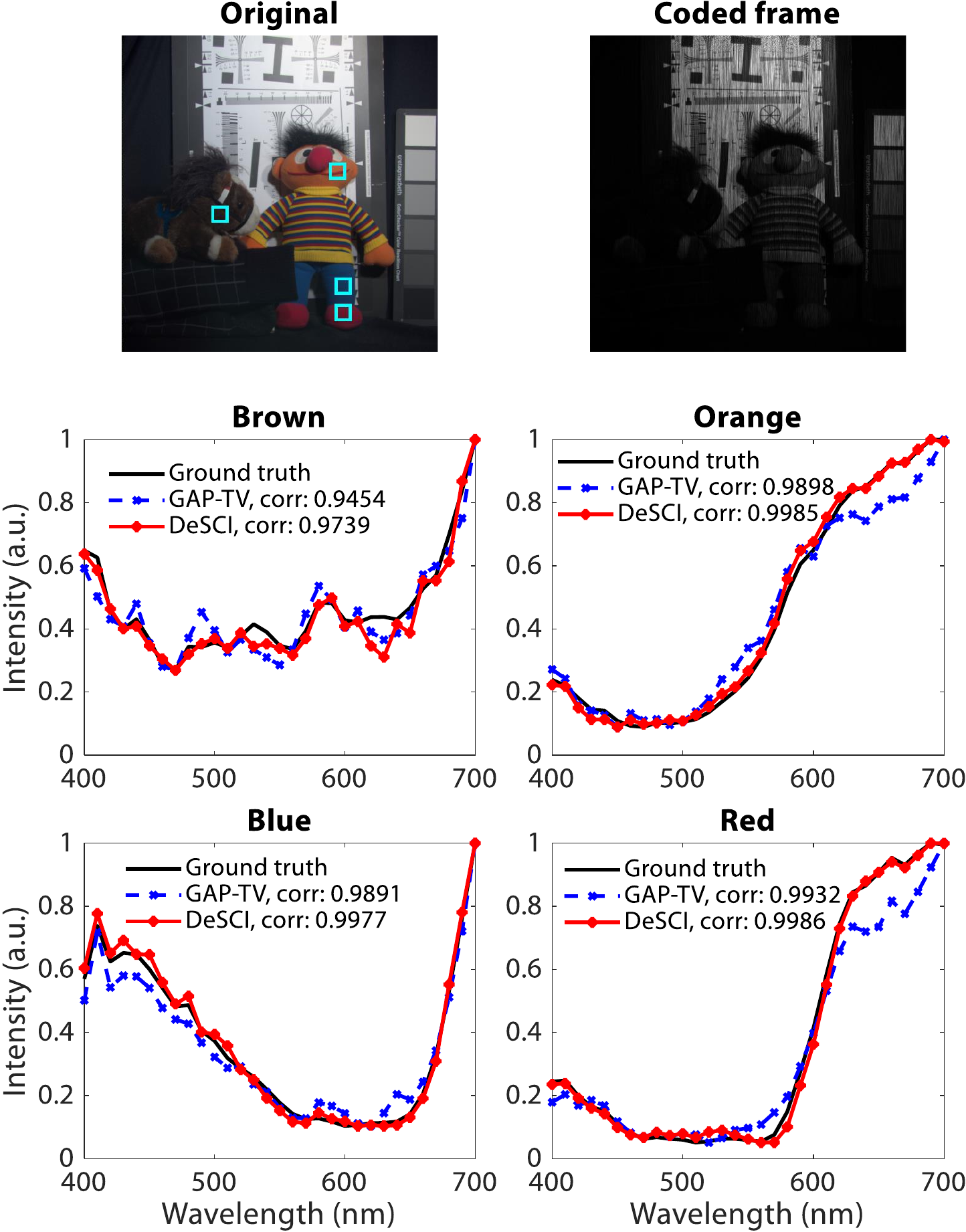}}
	\caption{Reconstructed spectra of \texttt{toy} hyperspectral data. A snapshot measurement encoding 31 spectral bands from simulated data is shown on the top-right. The \texttt{toy} data is from CAVE hyperspectral dataset. The original RGB image of the scene is shown on the top-left with a size of $512\times512$ pixels. The spectra of four color regions are shown on the middle and bottom rows. The correlation of the reconstructed spectra and the ground truth is shown in the legends.}
	\label{sfig:sim_hsi_toy_spectra}
\end{figure}
\begin{figure}
	\centering
	{\includegraphics[width=1\columnwidth]{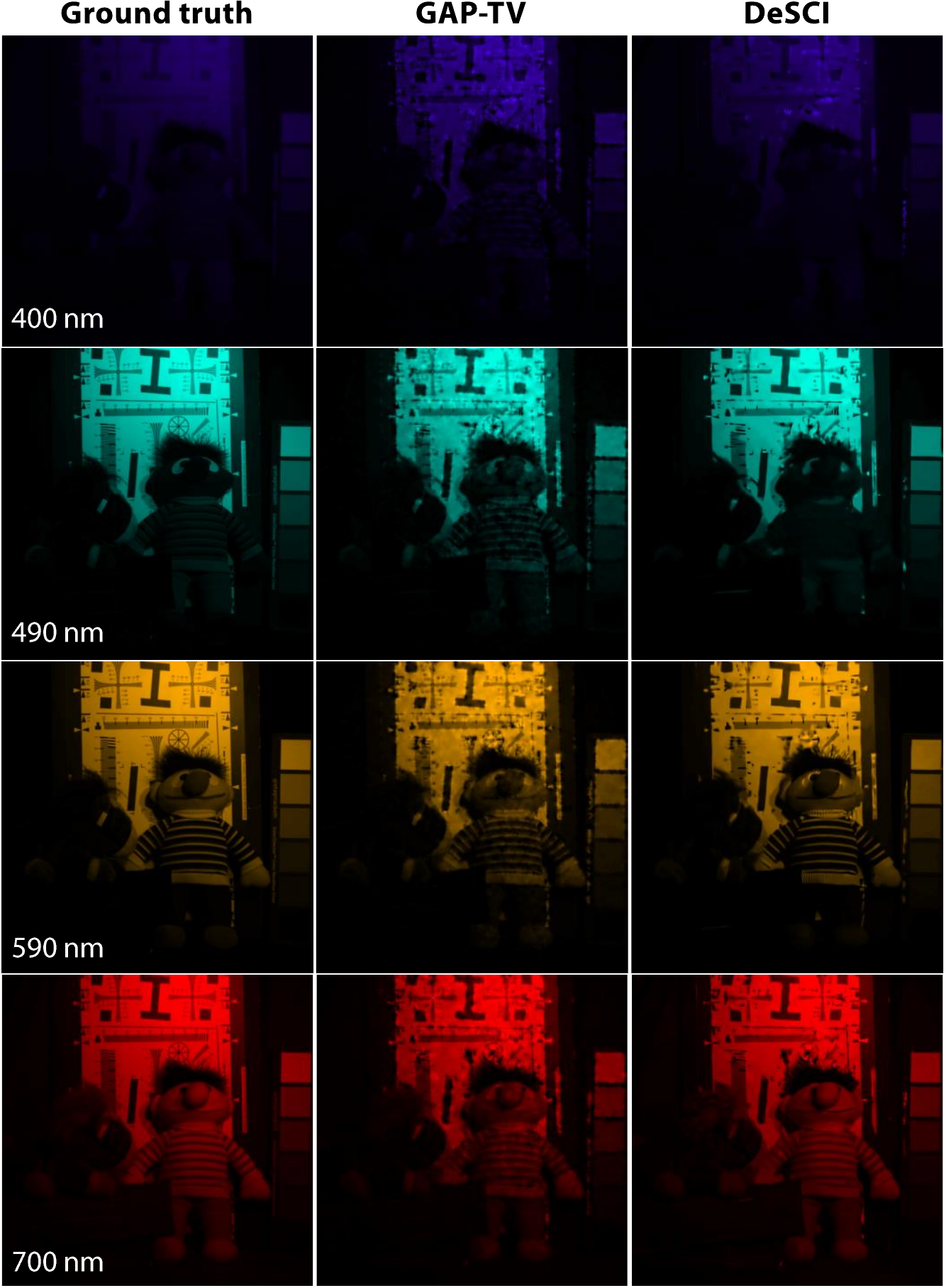}}
	\caption{Reconstructed frames of simulated \texttt{toy} hyperspectral data.}
	\label{sfig:sim_hsi_toy_frames}
\end{figure}
{\bf{\texttt{Toy} simulated hyperspectral data.}} The \texttt{toy} hyperspectral images consists of 31 spectral bands with each of size $512\times512$ pixels. The results of the reconstructed spectra of simulated \texttt{toy} hyperspectral data and exemplar frames are shown in Fig.~\ref{sfig:sim_hsi_toy_spectra} and Fig.~\ref{sfig:sim_hsi_toy_frames}, respectively. The averaged results of PSNR and SSIM are shown in Table~\ref{Tab:results_sim_hsi}. It can be seen clearly that DeSCI provides better reconstructed spectra than GAP-TV, as shown in the legends of Fig.~\ref{sfig:sim_hsi_toy_spectra}. Besides, the reconstructed frames of DeSCI preserves fine details of the \texttt{toy} dataset, for example the resolution targets and the fringe on the clothes in Fig.~\ref{sfig:sim_hsi_toy_frames} and gets higher PSNR and SSIM, as shown in Table~\ref{Tab:results_sim_hsi}.

\begin{figure*}
	\centering
	{\includegraphics[width=2\columnwidth]{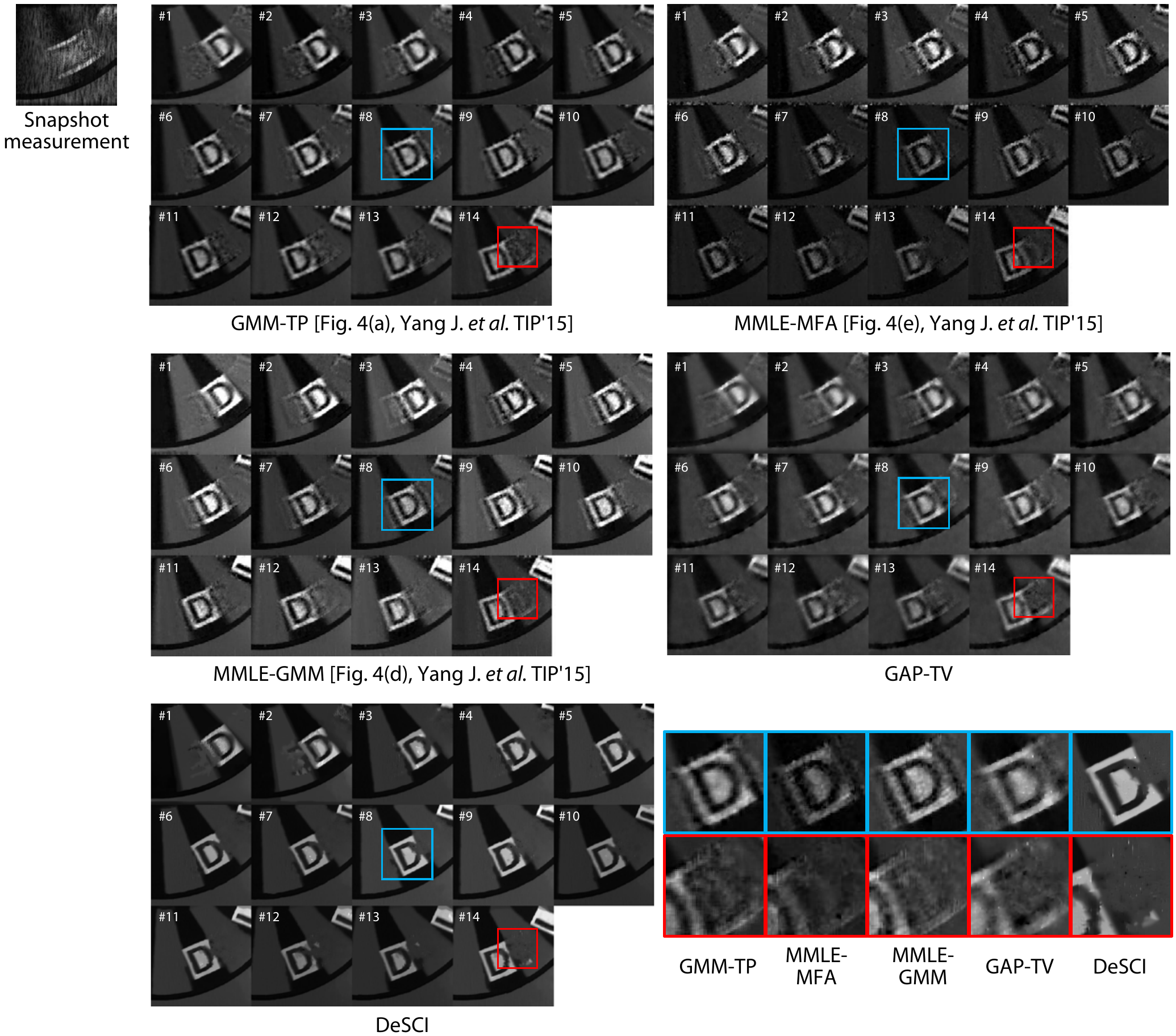}}
	\caption{Real data: Performance comparison on \texttt{chopper wheel} high-speed video reconstruction. A snapshot measurement encoding $14$ frames captured from the real SCI system~\cite{Patrick13OE} is shown on the top-left. Four algorithms (GMM-TP, MMLE-GMM, MMLE-GMM and GAP-TV) are compared with the proposed DeSCI algorithm. Close-ups of the letter `D' (in blue) and motion blur (in red) are shown on the bottom-right.}
	\label{sfig:real_chopperwheel}
\end{figure*}
\section{Real data results \label{Sec:Real_Results}}
In this section, we demonstrate the efficacy of the proposed DeSCI algorithm with real data captured from various SCI systems. Both grayscale and color video data and hyperspectral images are used for reconstruction and DeSCI provides significantly better reconstruction results.

\subsection{Grayscale high-speed video  \label{Sec:real_video_gray}} 
\label{sub:grayscale_video}
The grayscale high-speed video data are captured by the coded aperture compressive temporal imaging (CACTI) system~\cite{Patrick13OE}. A snapshot measurement of size $256\times256$ pixels encodes $14$ frames of the same size. The same data is used in two GMM papers~\cite{Yang14GMM,Yang14GMMonline}. We compare the results of DeSCI with three GMM based algorithms (GMM-TP, MMLE-GMM and MMLE-MFA) and re-use the figures\footnote{We re-use the figures in ~\cite{Yang14GMMonline} for two reasons. Firstly, we use exactly the same dataset as the GMM paper~\cite{Yang14GMMonline}. Secondly, only the pre-trained model of $B=8$ is provided by the authors~\cite{Yang14GMM}. However, for real data, $B=14$ for the grayscale high-speed video and $B=22$ for the color high-speed video.} from~\cite{Yang14GMMonline}. 

\textbf{\texttt{Chopper wheel} grayscale high-speed video.}  
The results of the \texttt{chopper wheel} dataset are shown in Fig.~\ref{sfig:real_chopperwheel}. It can be seen clearly that all other leading algorithms suffer from motion blur artifacts (see letter `D' and the corresponding motion blur in each frame and the close-ups in Fig.~\ref{sfig:real_chopperwheel}). However, DeSCI reserves both fine details and large-scale sharp edges. 

\begin{figure*}
	\centering
	{\includegraphics[width=2\columnwidth]{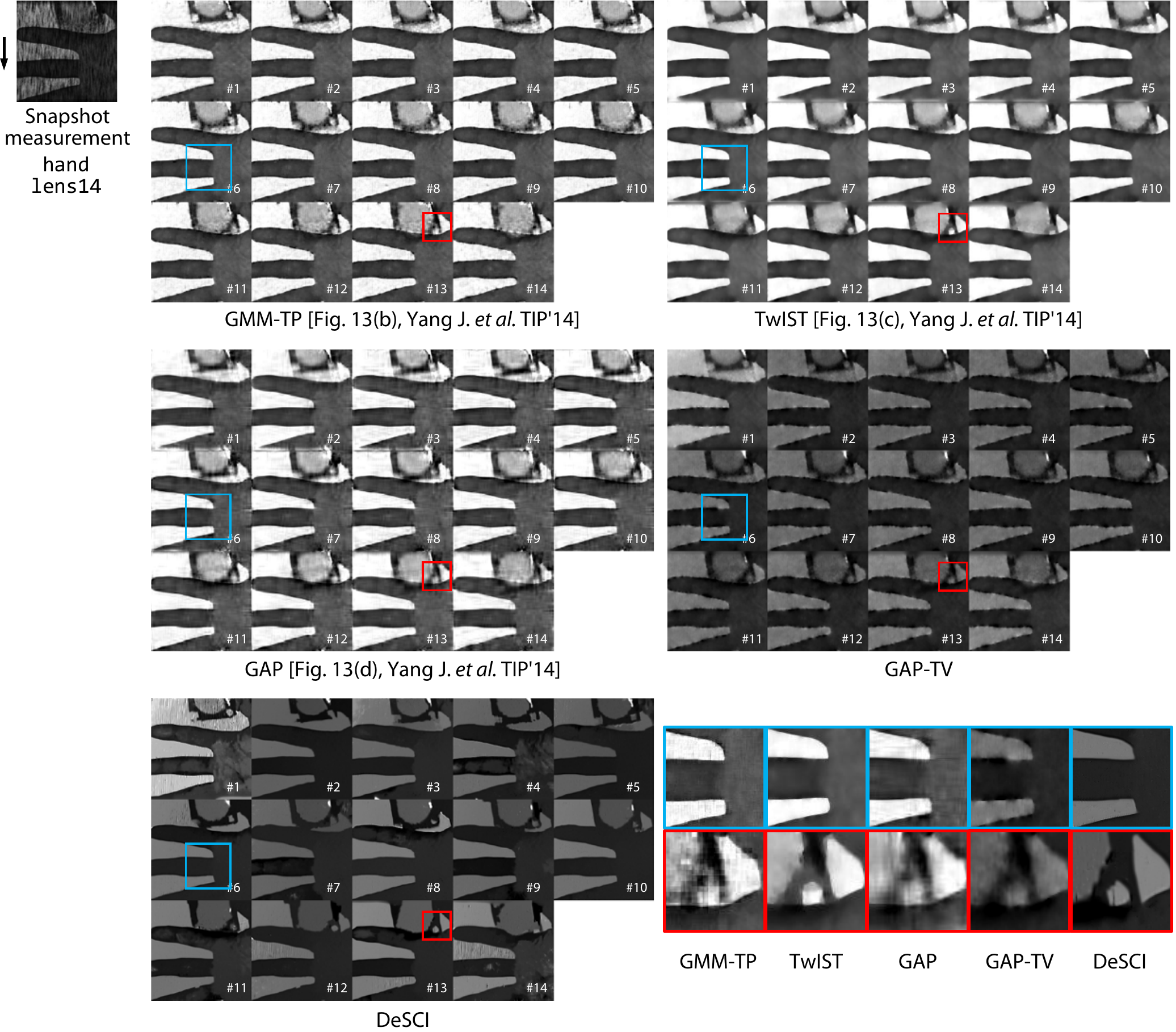}}
	\caption{Real data: Performance comparison on \texttt{hand lens} high-speed video reconstruction. A snapshot measurement encoding $14$ frames captured from the real SCI system~\cite{Patrick13OE} is shown on the top-left. Four algorithms (GMM-TP, TwIST, GAP and GAP-TV) are compared with the proposed DeSCI algorithm. Close-ups of the hand (in blue) and the lens (in red) are shown on the bottom-right.}
	\label{sfig:real_handlens}
\end{figure*}

\textbf{\texttt{Hand lens} grayscale high-speed video.}
Similar results of the \texttt{hand lens} dataset are shown in Fig.~\ref{sfig:real_handlens}. DeSCI reserves the clear background of the motionless hand (close-ups of the hand) and sharp edges of the moving lens (close-ups of the lens).

\begin{figure*}
	\centering
	{\includegraphics[width=2\columnwidth]{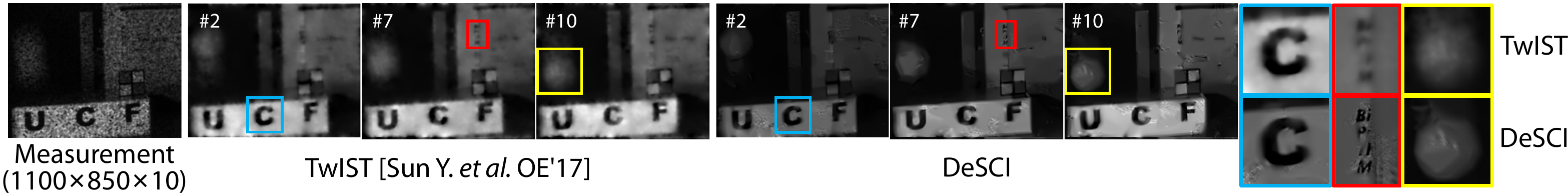}}
	\caption{Real data:  Performance comparison on \texttt{UCF} high-speed video reconstruction (raw data and code from~\cite{Sun17OE}). Close-ups shown on right.}
	\label{sfig:real_ucf}
\end{figure*}

\textbf{\texttt{UCF} grayscale high-speed video.}
DeSCI can also boost the performance of other snapshot high-speed compressive imaging systems. A snapshot measurement of size $1100\times850$ pixels encodes 10 frames of the same size from~\cite{Sun17OE}. We compare DeSCI and TwIST used in~\cite{Sun17OE} and results are shown in Fig.~\ref{sfig:real_ucf}. Three frames out of 10 total frames are shown here. DeSCI not only preserves sharp edges of the scene (close-ups of the `C' character and the moving ball), but also resolves fine details of the background. As shown in the middle close-ups, the characters of the book can be seen clearly in DeSCI reconstruction. However, TwIST blurs these details and no characters on the book can be identified. 


\subsection{Color high-speed video  \label{Sec:real_video_color}} 
\label{sub:color_video}
The color high-speed video data are from the high-speed motion, color and depth system~\cite{Yuan14CVPR}. A snapshot Bayer RGB measurement of size $512\times512$ pixels encodes $22$ frames of the same size. Since we do not have the pre-trained GMM model of $B=22$, we do not compare our algorithm DeSCI with three GMM based algorithms. As discussed in Sec.~\ref{sub:grayscale_video}, GMM based algorithms perform similar to GAP-TV \wrt motion blur artifacts. 

\begin{figure*}
	\centering
	{\includegraphics[width=2\columnwidth]{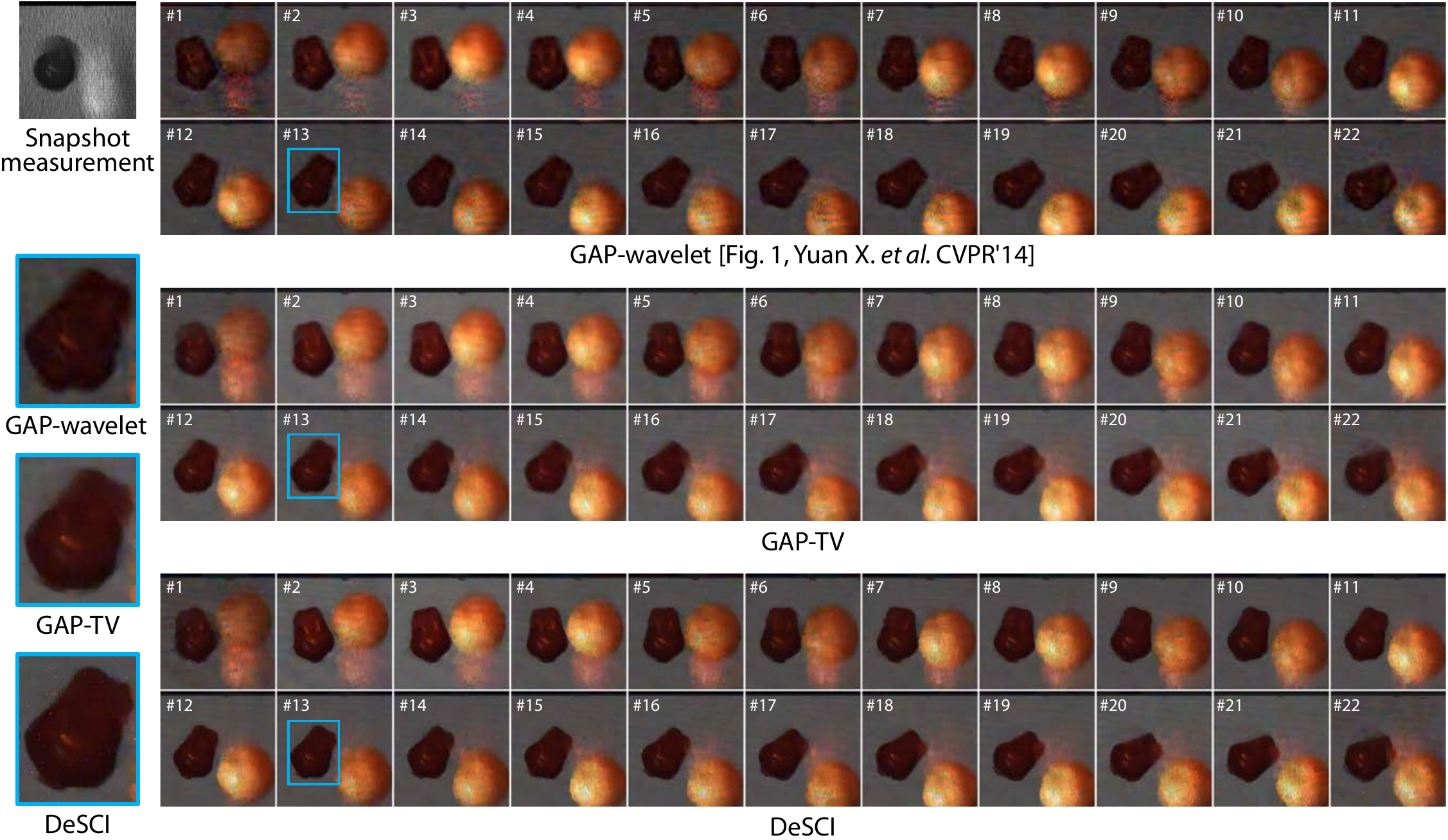}}
	\caption{Real data: Performance comparison on \texttt{triball} high-speed color video reconstruction. A snapshot Bayer RGB measurement encoding $22$ frames captured from the real SCI system~\cite{Yuan14CVPR} is shown on the top-left. A close-up of the dark red ball is shown on the left. GAP-TV is compared with the proposed DeSCI algorithm. }
	\label{sfig:real_triball}
\end{figure*}

\textbf{\texttt{Triball} color high-speed video.}
The results of the \texttt{triball} dataset are shown in Fig.~\ref{sfig:real_triball}. It can be seen clearly that the DeSCI results eliminate the motion blur artifacts compared with the GAP-TV results (see the close-up of the dark red ball in Fig.~\ref{sfig:real_triball}). Comparing with the GAP-wavelet results shown in~\cite{Yuan14CVPR}, our DeSCI results recovered a much smoother background. The orange ball still suffers from motion blur artifacts because its moving direction keeps the same as the shifting direction of the mask in all frames, and shift-mask-based SCI systems fail to resolve motions with the same direction as the shifting mask.

\begin{figure*}
	\centering
	{\includegraphics[width=2\columnwidth]{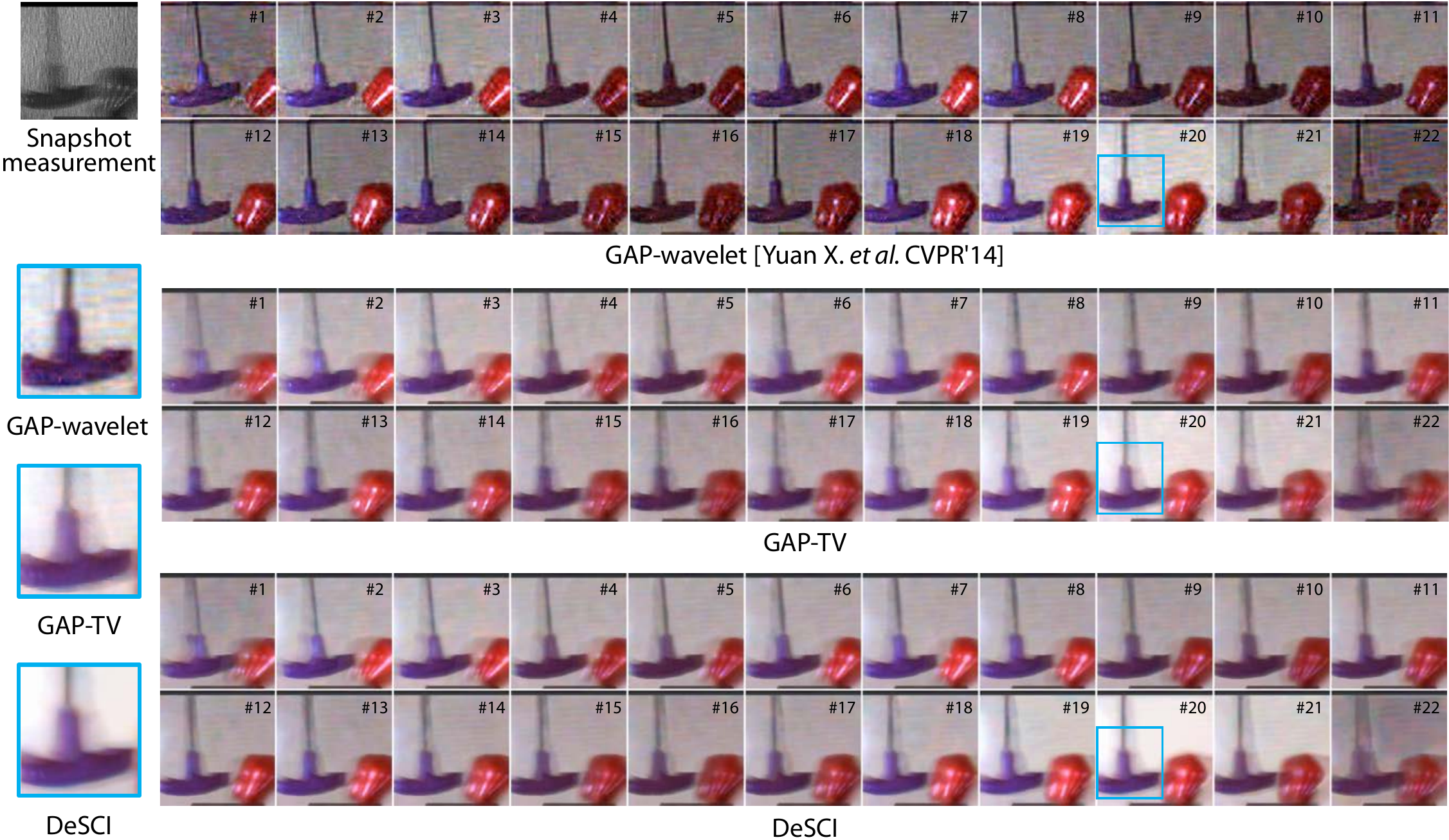}}
	\caption{Real data: Performance comparison on \texttt{hammer} high-speed color video reconstruction. Two snapshot Bayer RGB measurements encoding $22$ frames captured from the real SCI system~\cite{Yuan14CVPR} is shown on the top-left. A close-up of the hammer is shown on the left. GAP-TV is compared with the proposed DeSCI algorithm. }
	\label{sfig:real_hammer}
\end{figure*}

\textbf{\texttt{Hammer} color high-speed video.}
The results of the \texttt{hammer} dataset are shown in Fig.~\ref{sfig:real_hammer}. It can be seen clearly that the DeSCI results recovered a much smoother background compared with GAP-wavelet used in~\cite{Yuan14CVPR} and sharper edges of the hammer compared with GAP-TV (see the close-ups of the hammer in Fig.~\ref{sfig:real_hammer}).

\subsection{Hyperspectral image data  \label{Sec:real_hyper}} 
\label{sub:real_hyperspectral_data}
We apply DeSCI on real snapshot-hyperspectral compressive imaging data and show that DeSCI provides significantly better results by preserving the spectra and reducing artifacts induced by compressed measurements. 
The real hyperspectral data is from the CASSI system~\cite{Gehm07}. 

\begin{figure}
	\centering
	{\includegraphics[width=1\columnwidth]{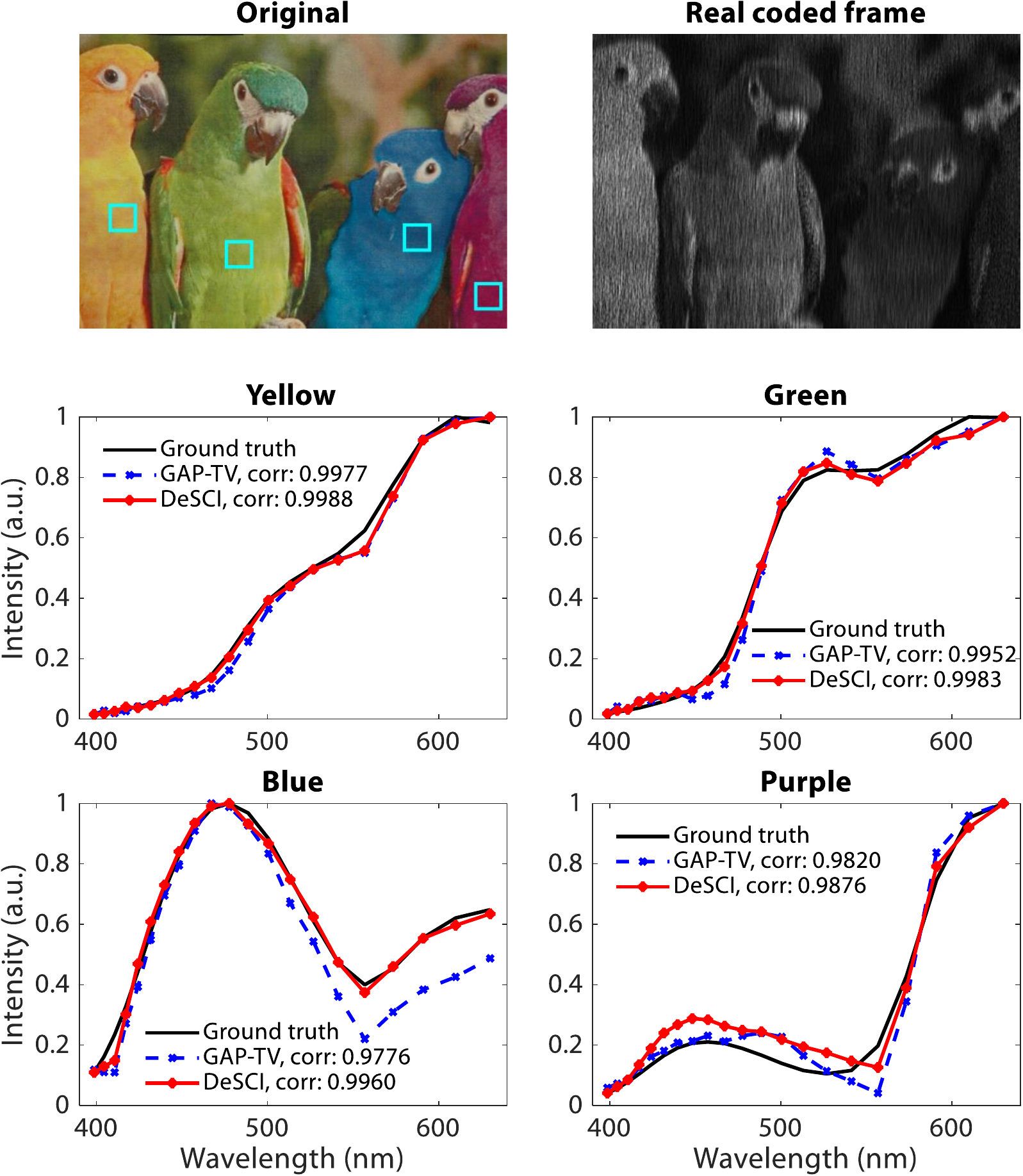}}
	\caption{Real data: Reconstructed spectra of real \texttt{bird} hyperspectral data. A snapshot measurement encoding 24 spectral bands captured from real CASSI system~\cite{Gehm07} is shown on the top-right. The original RGB image of the scene is shown on the top-left with a size of $1021\times703$ pixels. The spectra of four birds are shown on the middle and bottom rows. The correlation of the reconstructed spectra and the ground truth is shown in the legends.}
	\label{sfig:real_hsi_bird_spectra}
\end{figure}

\begin{figure}
	\centering
	{\includegraphics[width=1\columnwidth]{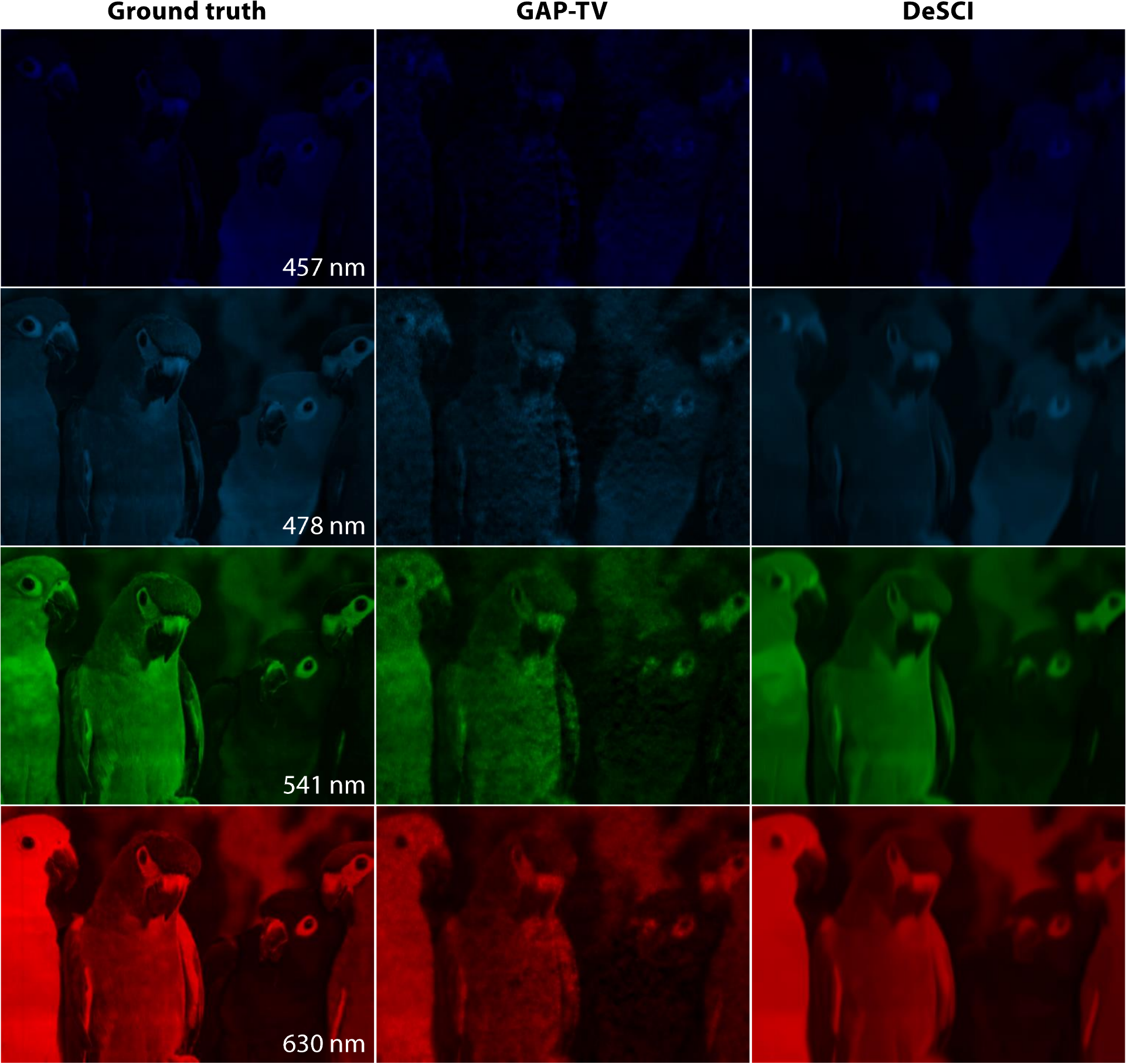}}
	\caption{Real data: Reconstructed frames of real \texttt{bird} hyperspectral data.} 
	\label{sfig:real_hsi_bird_frames}
\end{figure}

\textbf{ \texttt{Bird} hyperspectral data. } The \texttt{bird} data~\cite{Gehm07} consists of 24 spectral bands with each of size $1021\times703$ pixels. The reconstructed spectra and exemplar frames of the \texttt{bird} real hyperspectral data are shown in Fig.~\ref{sfig:real_hsi_bird_spectra} and Fig.~\ref{sfig:real_hsi_bird_frames}, respectively. We shift the reconstructed spectra two bands to keep align with optical calibration. So we only get 21 spectral bands shown in Fig.~\ref{sfig:real_hsi_bird_spectra}, compared with 24 bands shown in simulated reconstruction in Fig.~\ref{sfig:sim_hsi_bird_spectra}. It can be seen clearly that DeSCI preserves the spectra property of the scene with correlation over 0.99. Besides, the reconstructed frames of DeSCI are clearer than those of GAP-TV; the results of GAP-TV suffer from blurry artifacts resulted from coded measurements. We notice that there are some over-smooth phenomena existing in the DeSCI results on this real data in Fig.~\ref{sfig:real_hsi_bird_frames}, which might be due to the system noise in DeSCI reconstruction. 


\begin{figure}
	\centering
	{\includegraphics[width=1\columnwidth]{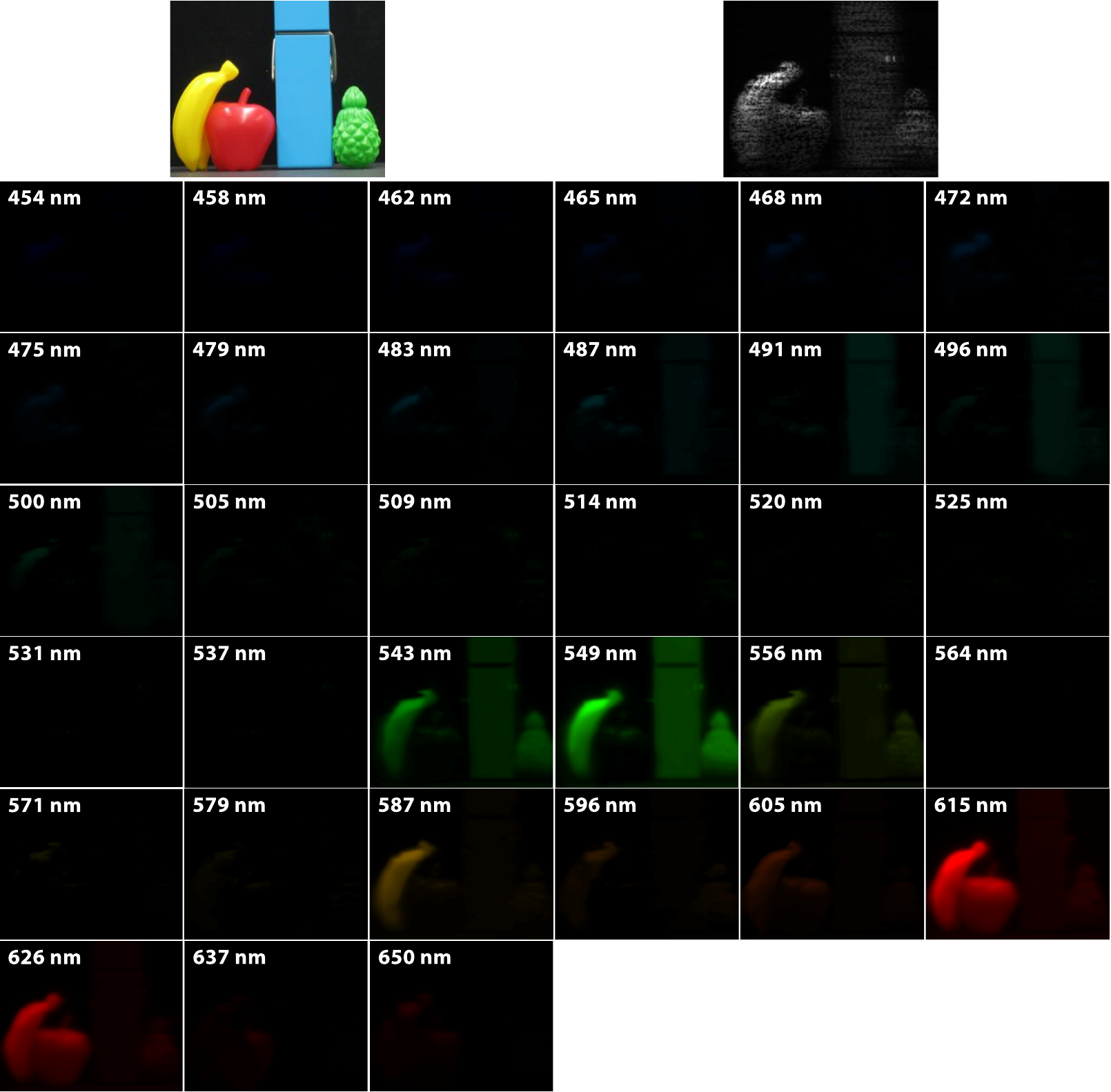}}
	\caption{Real data: Reconstructed frames of real \texttt{object} hyperspectral data. A snapshot measurement encoding 33 spectral bands captured from real CASSI system~\cite{Gehm07} is shown on the top-right. The original RGB image of the scene is shown on the top-left with a size of $256\times210$ pixels.}
	\label{sfig:real_hsi_object_all}
\end{figure}

\begin{figure}
	\centering
	{\includegraphics[width=\columnwidth]{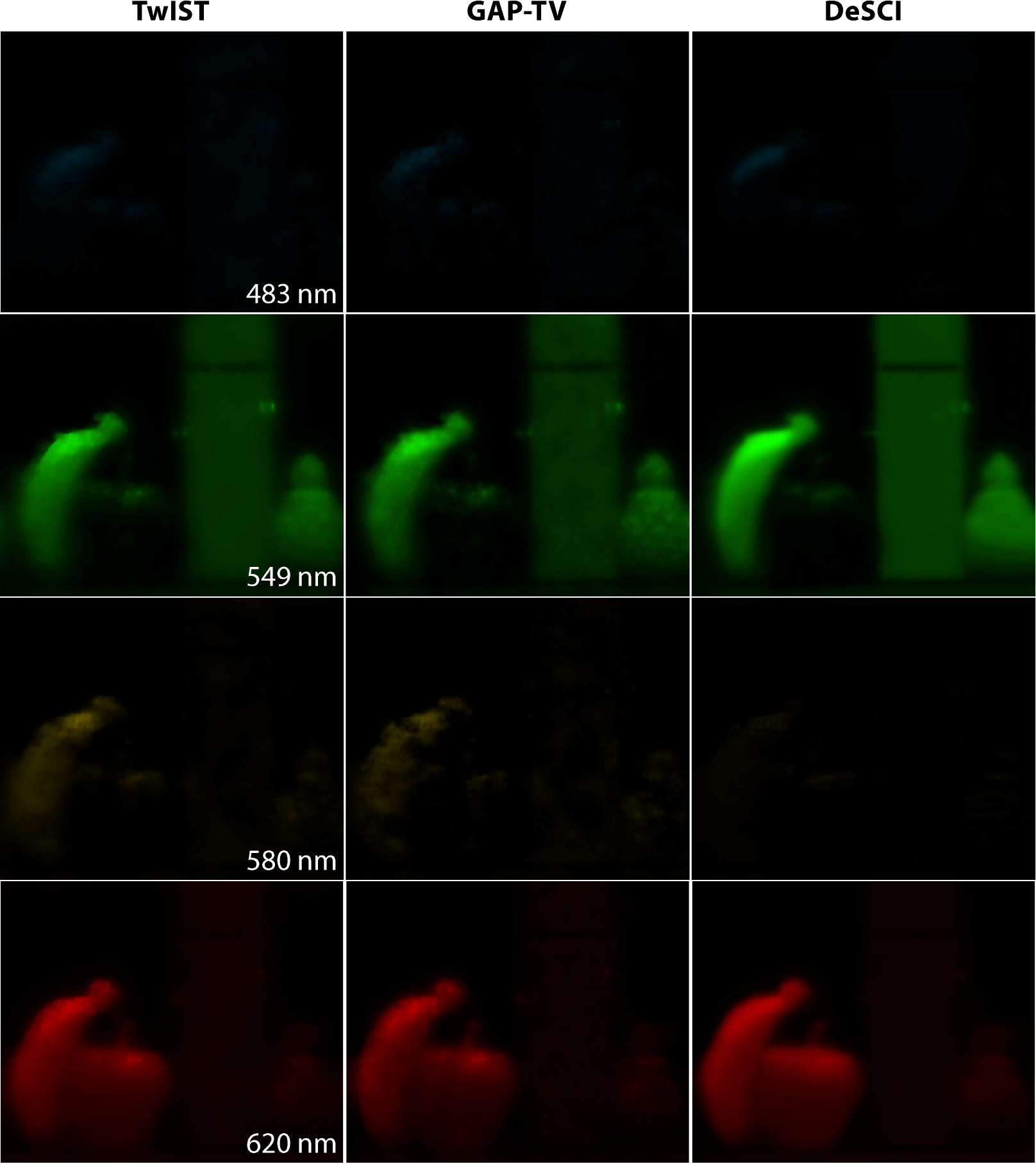}}
	\caption{Real data:  Reconstructed frames of real \texttt{object} hyperspectral data.}
	\label{sfig:real_hsi_object_frames}
\end{figure}

\textbf{ \texttt{Object} hyperspectral data. }  The \texttt{object} data~\cite{Gehm07} consists of 33 spectral bands with each of size $256\times210$ pixels. The reconstructed frames of DeSCI and comparison of exemplar frames with TwIST used in~\cite{Gehm07} and GAP-TV are shown in Fig.~\ref{sfig:real_hsi_object_all} and Fig.~\ref{sfig:real_hsi_object_frames}, respectively. Since there is no ground truth for the \texttt{object} data, we cannot compare the reconstructed spectra to the ground truth, nor the correlation. Our reconstructed frames are clear and free of blur. Each frame of a spectral band has a whole object or nothing, as shown in Fig.~\ref{sfig:real_hsi_object_all}. Because the plastic object with the same color in this scene are made of the same material, the whole object has the same spectral band. In contrast to the yellow band of the TwIST and GAP-TV results shown in Fig.~\ref{sfig:real_hsi_object_frames}, DeSCI sees no fraction of the banana object. DeSCI reduces the mask artifacts significantly induced by compressed measurements.

\section{Concluding remarks \label{Sec:Dis}}
We have proposed a new algorithm to reconstruct high-speed video frames from a single measurement in snapshot compressive imaging systems, with videos and hyperspectral images as two exemplar applications.
The rank minimization approach is incorporated into the forward model of the snapshot compressive imaging system and a joint optimization problem is formulated.
An alternating minimization algorithm is developed to solve this joint model. 
Our proposed algorithm has investigated the nonlocal self-similarity in video (hyperspectral) frames, and has led to significant improvements over existing algorithms.
Extensive results of both simulation and real data have demonstrated the superiority of the proposed algorithm.     

 
Most recently, the deep learning algorithms have been used in CS inversion~\cite{Kulkarni_2016_CVPR,ADMMnet2016NIPS,Yuan18OE} and as mentioned in the introduction, deep learning has been employed for video CS in~\cite{Iliadis18DSPvideoCS,2016arXivVideoCS} (but with some constraints). 
While most of these algorithms tried to learn an end-to-end inversion network via convolutional neural networks (CNN), Yang {\em et al.}~\cite{ADMMnet2016NIPS} incorporates the CNN into the ADMM framework. This inspires one of our future work to integrate the proposed ADMM framework with the deep learning based denoising algorithms~\cite{Zhang17TIP_DeepNoise,Zhang17SPM_deepdenoise}. 

Our proposed algorithm can be used in other snapshot (or multiple shots) compressive imaging systems, for example, the depth compressive imaging~\cite{Llull15Optica}, polarization compressive imaging~\cite{Tsai15OE}, X-ray compressive imaging~\cite{Wang15SIAM_xray,Huang15ICASSP_xRay,Huang15ICASSP_multiscale} and three-dimensional high-speed compressive imaging~\cite{Sun16OE} systems. This will be another direction of our future work.
We believe our results  will encourage the researchers and engineers
to pursue further in compressive imaging on real applications.

\section*{Acknowledgments}
The authors would like to thank Dr. Patrick Llull for capturing the real video data of CACTI, Dr. Tsung-Han Tsai for capturing the real hyperspectral image data of CASSI,  and Mr. Yangyang Sun and Dr. Shuo Pang for providing the UCF video data. This work was supported by the National Natural Science Foundation of China (grant Nos. 61327902, 61722110, 61627804, and 61631009).

\bibliographystyle{IEEEtran}
\bibliography{IEEEabrv,DeSCI_TPAMI_arXiv}
\end{document}